\documentclass[onecolumn,10pt, cleanfoot]{asme2ej}

\usepackage{graphicx} 
\usepackage{xcolor}
\usepackage{amsmath}
\usepackage{bbold}
\usepackage{bm} 
\usepackage{amsfonts}
\usepackage{amsmath}
\usepackage{mathrsfs}
\usepackage{bm}
\usepackage{amsfonts}
\usepackage{color}
\usepackage{xcolor}
\usepackage{verbatim}
\usepackage{subfigure}
\usepackage{cleveref}
\usepackage{algorithm}
\usepackage{algpseudocode}

\newcommand{\bx}{\mathbf{x}}
\newcommand{\bX}{\mathbf{X}}
\newcommand{\bp}{\mathbf{p}}
\newcommand{\bq}{\mathbf{q}}

\newcommand{\by}{\mathbf{y}}

\newcommand{\du}{d\mathcal{U}}

\newcommand{\balpha}{\boldsymbol{\alpha}}

\newcommand{\btheta}{\boldsymbol{\theta}}

\newcommand{\bphi}{\boldsymbol{\phi}}

\newcommand{\R}{\mathbb{R}}

\newcommand{\mc}{\mathcal}
\newcommand{\calD}{\mc{D}}

\newcommand{\calO}{\mc{O}}
\newcommand{\calS}{\mc{S}}

\newcommand{\order}[1]{\calO({#1})}

\newcommand{\sref}[1]{Sec.~\ref{sec:#1}}
\newcommand{\qref}[1]{Eq.~(\ref{eqn:#1})}

\newcommand{\fref}[1]{Fig.~\ref{fig:#1}}

\title{Advances in Bayesian Probabilistic Modeling for Industrial Applications}

\author{Sayan Ghosh \thanks{Address all correspondence to this author at sayan.ghosh1@ge.com }\\
	Lead Engineer\\
	{\tensfb Piyush Pandita, Steven Atkinson, Waad Subber, Yiming Zhang}\\
	Research Engineer\\
	{\tensfb Natarajan Chennimalai Kumar, Suryarghya Chakrabarti}\\
	Senior Engineer\\
	{\tensfb Liping Wang}\\
	Technology Manager\\
    \affiliation{
	Probabilistic Design and Optimization Lab\\
	GE Research\\
	Niskayuna, New York\\
    }	
}

\begin{document}

\maketitle    

\begin{abstract}
{\it Industrial  applications frequently pose a notorious challenge for state-of-the-art methods in the contexts of optimization, designing experiments and modeling unknown physical response. This problem is aggravated by limited availability of clean data, uncertainty in available physics-based models and additional logistic and computational expense associated with experiments. In such a scenario, Bayesian methods have played an impactful role in alleviating the aforementioned obstacles by quantifying uncertainty of different types under limited resources. These methods, usually deployed as a framework, allows decision makers to make informed choices under uncertainty while being able to incorporate information on the the fly, usually in the form of data, from multiple sources while being consistent with the physical intuition about the problem. This is a major advantage that Bayesian methods bring to fruition especially in the industrial context. This paper is a compendium of the Bayesian modeling methodology that is being consistently  developed at GE Research. 
The methodology, called GE's Bayesian Hybrid Modeling (GEBHM), is a probabilistic modeling method, based on the Kennedy and O'Hagan framework, that has been continuously scaled-up and industrialized over several years.
In this work, we explain the various advancements in GEBHM's methods and demonstrate their impact on several challenging industrial problems. }
\end{abstract}

%

\section{Introduction}

Reliability of industrial applications necessitates physical experiments of complex engineering systems. 
In conjunction with their cost, measurements  generated from these complex physical tests can be limited, or even sometimes infeasible, in practice due to lack of accessibility and repeatability. 
Advances in numerical algorithms and the availability of high-performance computing platforms, on the other hand, facilitate solution of complex engineering systems with less cost, in general, than the physical experiments. 
For credible numerical simulations, computers models should be verified, validated and calibrated. 
The numerical simulations can only  be  {\it predictive simulations} once uncertainties due to model and input parameters are quantified~\cite{oden2010computer}.
In abstract sense, both the physical experiments and predictive computer models can be viewed as data generation mechanisms. 
As the complexity of the industrial applications increases, a unified framework linking experimental measurements and numerical simulations  is becoming more critical than ever. 

To this end, the Bayesian paradigm provides a rich probabilistic approach to combine information from  physical experiments and computer simulations in a way that permits engineers to continuously update their prior belief on the process under consideration~\cite{oden2010computer}. In Bayesian framework, a probabilistic meta-model can be constructed utilizing data from physical experiments as well as low and high fidelity  computer simulations. Optimally, design of computer experiments (DoE) is performed to guide the construction of such meta-models. The DoE algorithms seek to balance between the cost and accuracy of the computer simulations~\cite{ghosh2019strategy,kristensen2019industrial}. Industrial-scale computer simulations often have large input dimensions. The input space may include control variables that are set to control the desired process and physical parameters reflecting the media of operation~\cite{santner2003design}. Therefore, sensitivity analysis can estimate the relative influence of each input parameter on the variability of the prediction. The most influential parameters are then calibrated under uncertainty for accurate representation of important aspects of the physical process~\cite{hill2000methods}. Once calibrated, uncertainties due to modeling errors and data are propagated  to predict the quantities of interest with confidence bounds.

For the Bayesian framework, building surrogate models utilizing both high and low-fidelity  simulations as well as experimental data are often based on Gaussian Processes (GP) which has become an indispensable tool in the Bayesian paradigm~\cite{Kennedy2001}.  In this direction, GP surrogates have been applied for various engineering problems~\cite{forrester2009recent,wang2007review}, such as optimization of composite laminates\cite{zhang2016approaches}, structural optimization\cite{hao2012surrogate}, combustion synthesis of composite materials~\cite{shabouei2019chemo}, design applications~\cite{zhang2013crashworthiness,chaudhuri2015experimental} and reliability analysis~\cite{hu2016single}, and additive manufacturing~\cite{tapia2016prediction}, to name a few.

\begin{figure}[t]
	\begin{center}
		\includegraphics[scale=0.35]{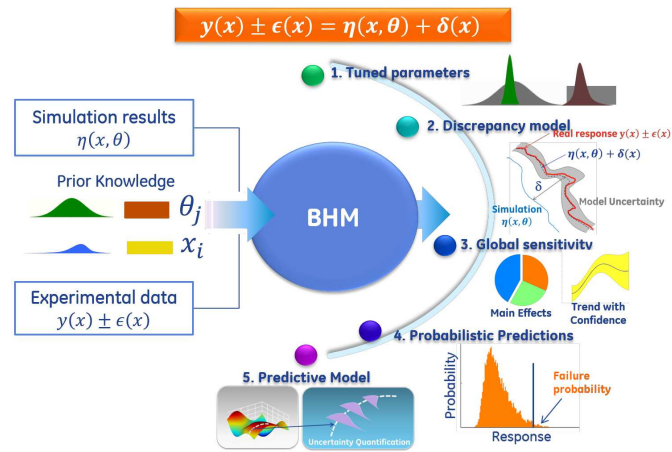}
	\end{center}
	\caption{GE's Bayesian Hybrid Modeling  (GEBHM) framework.}
	
	\label{fig:BHM} 
\end{figure}

The Kennedy and O'Hagan approach~\cite{Kennedy2001} can be viewed as one of the well established unified frameworks for Bayesian calibration, meta-modeling utilizing both simulation model and observed data. A computer  implementation (named \textit{gpmsa}) to this approach was released by Los Alamos National Lab ~\cite{Higdon2008,gattiker2016gaussian}. Deploying \textit{gpmsa} to challenging industrial problems is not a straightforward task, as many technical challenges exist, such as the curse of dimensionality, model-identifiability, transient problems with multiple outputs, lack of data and extrapolation, multiple sets of experimental data, etc. To overcome these challenges, we at GE Research have built the GE Bayesian Hybrid Modeling (GEBHM) framework (shown in Fig.\ \ref{fig:BHM}) to carry out Bayesian analysis for industrial problems. The core components of GEBHM are based on \textit{gpmsa}, but with several significant enhancements on top that are required for robust employment to many industrial problems.

Presented in Fig. \ref{fig:BHM}, the distinguished features of GEBHM are: (1)  using full-Bayesian approach (2) an explicitly formulated model discrepancy (3) building the GP surrogates during the calibration and updating process (4) calibration of  dynamic models with multiple outputs (5) forward uncertainty quantification and propagation (6) model validation (7) real time visualization of the calibration process and outcomes (8) scaling to high dimensional problems (i.e., hundreds of parameters) (8) parallelization  on high-performance computing platforms. These capabilities of GEBHM have been successfully demonstrated on several challenging engineering applications, such as multi-stage engine Blade Row Models, transient model calibration , engine system-level thermal model calibration and validation, combined cycle plant (gas turbine and steam turbine), control model calibration, combustor global sensitivity, predictive emissions modeling and optimization, airfoil cooling system uncertainty quantification and model validation, engine cycle deck performance data matching, Airframe Digital Twin, and more.

In this work, we present the  GEBHM  framework and demonstrate its capability in tackling many industrial applications. In particular, Section~\ref{sec:gebhm_calibration} briefly reviews the theoretical background of the framework and an application on structural dynamic problem. In Section~\ref{Applications} we show GEBHM applicability to transient problem, robust design, sensitivity analysis and portable GP, multi-source legacy data modeling, and parallel MCMC. Section~\ref{Future} we discuss some of our future enhancements to GEBHM.

\section{GE's Bayesian Hybrid Modeling (GEBHM)}

\label{sec:gebhm_calibration}

Before delving into the advancement, we will first  discuss the the Kennedy and O'Hagan theoretical framework.

In this section, we discuss the Kennedy and O'Hagan theoretical framework framework for data assimilation. In this framework, let $y$ denote the outputs from either a physical process or a high fidelity simulator, and $\bm{x}$ represents a design vector of $p$ dimension. The training  output $y_{i}$, at training \ input point $\bm{x}_i$ can be expressed as:
\begin{align}
\label{eqn:koh_y}
& y_{i}\pm\epsilon_{ij}(\bm{x}_i)=\eta(\bm{x}_i,\bm{\theta}^*) + \delta(\bm{x}_i),~ \mbox{for}\; \; i=1, \; \ldots\; ,n,  
\end{align}
where  $n$ is the number of training data points, $\eta$ is the simulation model, $\bm{\theta}^*$ are the true values of calibrated parameters, $\delta$ is the discrepancy model between the calibrated  model and experimentally observed data, and $\epsilon$ is noise in the observed data.
Generally, the data from the simulation model are available on a set of $m$ design and calibration parameter combination $\eta(\bm{x}_j,\bm{\theta}_j)$ for $  j=1, \; \ldots\; ,m$, which may not be  at the same setting of experimentally observed data.  
As proposed by Kennedy and O'Hagan \cite{Kennedy2001}, and as described by Higdon \textit{et al.} \cite{Higdon2008}, the simulation output and the model discrepancy are modeled as GP models. 
The simulator model ($\eta$) is approximated as a GP model using known simulator output at $m$ design locations  with a zero mean and covariance matrix given by,
\begin{align}
\Sigma_{ij}^{\eta}  = \frac{1}{\lambda_{\eta_z}} \exp \left( \beta_{\eta}  | \bm{x}^{sim}_i - \bm{x}^{sim}_j |^2 \right) + I \frac{1}{\lambda_{\eta_s}}~\mbox{for} \; \; i,j = 1,\dots, m,
\label{eq:sigma_eta}
\end{align}
\noindent where $\bm{x}^{sim}$ is the combined vector of design variables and calibration parameter $\left(\bm{x}^{sim}_j = (\bm{x}_j,\bm{\theta}_j)\right)$, the parameters $\lambda_{\eta_z}$ and $\lambda_{\eta_s}$ characterize the marginal data variance captured by the model and by the residuals, respectively, and $\beta_{\eta}$ characterizes the strength of dependence of the outputs on the design variables.
The squared exponential function ensures that the GP model is smooth and is infinitely differentiable. 
Similarly, the experimental observation is modeled as zero mean and covariance matrix given by,
\begin{align}
 \Sigma_{ij}^{y}  = \frac{1}{\lambda_{y_z}} \exp \left( \beta_{y}  | \bm{x}^{obs}_i - \bm{x}^{obs}_j |^2 \right) + I \frac{1}{\lambda_{y_s}}~ \mbox{for} \; \; i,j = 1,\dots,  n,
\label{eq:sigma_y}
\end{align}
where $\bm{x}^{obs}$ is the combined vector of observed design variables of experiments and calibration parameters $\left(\bm{x}^{obs}_i = (\bm{x}_i,\bm{\theta}_i^*)\right)$ and $\bm{\theta}_i^*$ is the prior values of calibration parameter. The parameters $\lambda_{y_z}$, $\lambda_{\eta_s}$, and $\beta_{y}$ are for observation GP model.
To represent the correlation between simulator output and experimental observations, the cross covariance matrix is given as

\begin{align}
& \Sigma_{ij}^{\eta y}  = \frac{1}{\lambda_{y_z}} \exp \left( \beta_{y}  | \bm{x}^{obs}_i - \bm{x}^{sim}_j |^2 \right) + I \frac{1}{\lambda_{y_s}}  \\
& \mbox{for} \; \; i = 1,\dots n, j = 1,\dots, m. \nonumber 
\end{align}
The discrepancy parameter $\delta$ is modeled as 
\begin{align}
\label{eq:sigma_delta}
& \Sigma_{ij}^{\delta}  = \frac{1}{\lambda_{\delta_z}} \exp \left( \beta_{\delta}  | \bm{x}_i - \bm{x}_j |^2 \right) \\
& \mbox{for} \; \;  i,j = 1,\dots, n \nonumber
\end{align}
\noindent where, $\lambda_{\delta_z}$ and $\beta_{\delta} $ are the parameters associated with discrepancy model GP. 

Instead of operating on the raw data from the two sources namely, a) experimental, and b) simulation, we scale the observations to a standard normal using the mean and standard deviation of the simulation data. This is done in order to remain consistent with the zero mean function of the GP based models for the simulation and the discrepancy terms. Another important feature of the data-processing phase is the transformation of the experimental normalized observed data into two parts: a) underlying function and, b) discrepancy term. The transformed simulation data is denoted by $y_{s}$.
The underlying function part from the transformed experimental data is denoted by $y_{o}$ and the discrepancy part by $y_{d}$.
This is done using a linear basis transformation, the details of which are more involved and outside the scope of the discussion here.
More interested readers are referred to Section 3.2  of \cite{gattiker2016gaussian} where the aforementioned data preprocessing steps are exhaustively discussed.
Intuitively, the $(y_{o}^T, y_{s}^T)$ and  $y_{d}^T$, both of which are assumed to be samples from multivariate \emph{zero-mean} Gaussian distributions, serve as the training data for the simulation and the discrepancy GP models, respectively.
 
The likelihood of combined data $z = (y_{o}^T, y_{s}^T, y_{d}^T)$ is then given as follows:
\begin{equation}
\label{eqn:likelihood}
\mathbb{L}(D|\lambda_{\eta_z}, \beta_{\eta}, \lambda_{\eta_s},\lambda_{\delta_z},\beta_{\delta}) = \frac{1}{|\Sigma|^{1/2}} \exp{\left( - \frac{1}{2} D^T \Sigma ^{-1} D \right)}
\end{equation}
\noindent where $D = (y_{o}^T, y_{s}^T, y_{d}^T)$ and $\Sigma = \begin{pmatrix}
\Sigma^{y_{d}}  & 0 &  0 \\ 0 & \Sigma^{y_{o}} & \Sigma^{y_{o}y_{s}} \\ 0 &  \Sigma^{y_{s}y_{o}} & \Sigma^{y_{s}}
\end{pmatrix}$
\noindent For multiple outputs $y$, additional processing such as Singular Value Decomposition (SVD), support kernels for computing $\delta$, etc. are required for which readers can refer to  \cite{Higdon2008}.
The posterior distribution of all the hyper-parameters is given by 
\begin{align}
& \pi (\lambda_{\eta_z}, \beta_{\eta}, \lambda_{\eta_s},\lambda_{\delta_z},\beta_{\delta} | D) = \\ & \mathbb{L}(D|\lambda_{\eta_z}, \beta_{\eta}, \lambda_{\eta_s},\lambda_{\delta_z},\beta_{\delta}) \pi(\lambda_{\eta_z}) \pi(\beta_{\eta}) \pi(\lambda_{\eta_s}) \pi(\lambda_{\delta_z}) \pi(\beta_{\delta}) \nonumber
\end{align}
\noindent where $\pi(.)$ on the right-hand side of the equation is the prior distribution of the parameters, where all the hyperparameters are assumed to be independent. The target posterior distribution is evaluated using Metropolis-Hastings within the Gibbs algorithm with univariate proposal distributions for the MCMC posterior updates \cite{Wang2011, Subram2012, Kennedy2001, Higdon2008}.
The covariance matrices are updated with current realizations of the hyperparameters at every MCMC step.

It is important to emphasize  a fundamental principle of the mathematical formulation in the calibration framework, which is based on the so-called Kennedy O'Hagan formulation~\cite{kennedy2001bayesian,Higdon2008}. This formulation models the simulation model, $\eta(\bm{x}, \bm{\theta})$, as a GP regression model using all the available data i.e. simulation data and observation data. 
This is why the likelihood formulation, in Eq.~\ref{eqn:likelihood}, has a cross-covariance between the observed data and simulation data.  This approach is different from other model calibration methods that directly calibrate the known physical parameters of the simulation model using the observed data.


For a test input point matrix $\bm{X}_*$, the predicted mean and standard deviation for the $g$\textsuperscript{th} output is given as,
\begin{align}
\label{eq:pred_mean_cov}
& \mu(\bm{X}_*) = \Sigma(\bm{X}_*, \bm{X})\Sigma(\bm{X}, \bm{X})^{-1}y   \\
& \Sigma(\bm{X}_*, \bm{X}_*) = \Sigma(\bm{X}_*, \bm{X}_*) - \Sigma(\bm{X}_*, \bm{X}) \Sigma(\bm{X}, \bm{X}) ^{-1}\Sigma(\bm{X}, \bm{X}_*) \nonumber
\end{align} 
\noindent where $\bm{X}$ is the observed data matrix and $y$ is the vector the observed output used for training the model.

\subsubsection*{Application on Structural Dynamics Problem}
\label{Structural_Dynamics}
\begin{figure}[h]
	\begin{center}
		\includegraphics[scale=0.4]{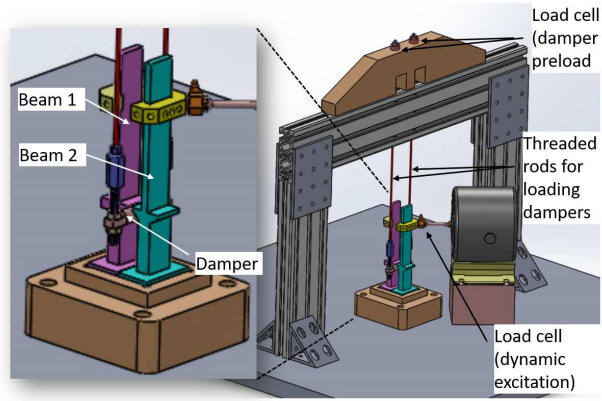}
	\end{center}
	\caption{Vibration test setup for structural dynamic problem}
	\label{fig:benchrig} 
\end{figure}

This section presents a new demonstration for the GEBHM calibration capability. In particular, we consider a physical experiment to model the underplatform dampers used widely in turbine blades and investigate the damping effectiveness of different damper geometries over a wide range of contact preload and dynamic excitation levels. Fig.~\ref{fig:benchrig} illustrates the experiment setup, which consists of two rectangular beams placed adjacent to each other. The beams have a platform at roughly one third of the beam length from the fixed end. A vibration damper is placed between them such that it contacts the underside of the beam platforms. The damper is held in place by two threaded rods which are attached to surrounding stationary frame. The damper can be loaded up against the platforms by pretensioning the rods and this load is measured by loadcells between the pretensioning nuts and the stationary frame. Dynamic excitation is provided to one of blades using an electrodynamic shaker.  Analytical simulation of the experiment is performed by generating a Craig-Bampton type reduced order model from the finite element model of the test setup shown in Fig.~\ref{fig:benchrig} and performing a nonlinear vibration analysis using a multi-harmonic balance hybrid frequency time domain solver similar to that described by Poudou et al. \cite{poudou2003hybrid} This helps improve the solution times by several orders of magnitude over the traditional approach of using a full FEA model with a transient time stepping analysis.

The problem has two design variables: 1) the static damper preload and 2) the dynamic excitation. Furthermore, there are  two unknown calibration parameters that are 1) the friction coefficient $(\mu)$ and 2) backround structural damping which is applied as purely stiffness proportional damping with the proportionality constant $\beta$.  The design variables preload and excitation are denoted by $x_1$ and $x_2$, respectively.

The friction coefficient $\mu$ and the background damping $\beta$ are the parameters being inferred. Both the test and simulation provide response as a function of excitation frequency. The frequency responses are parameterized by two parameters. The maximum response and the damping expressed as Q-factor. The former captures the height of the response peaks while the latter captures the width of the frequency response function peaks. We make use of the test data obtained for 24 designs  and simulation data at 30 designs. \fref{calibration_vibration}. shows the posterior densities of the calibration parameters $\mu$ and $\beta$ obtained using GEBHM calibration framework. Whereby, GEBHM estimates the posterior distributions by fusing information from the two sources and using the fully-Bayesian MCMC discussed in \sref{gebhm_calibration}.

\begin{figure}[h]
  \subfigure[]{\includegraphics[width=0.5\textwidth]{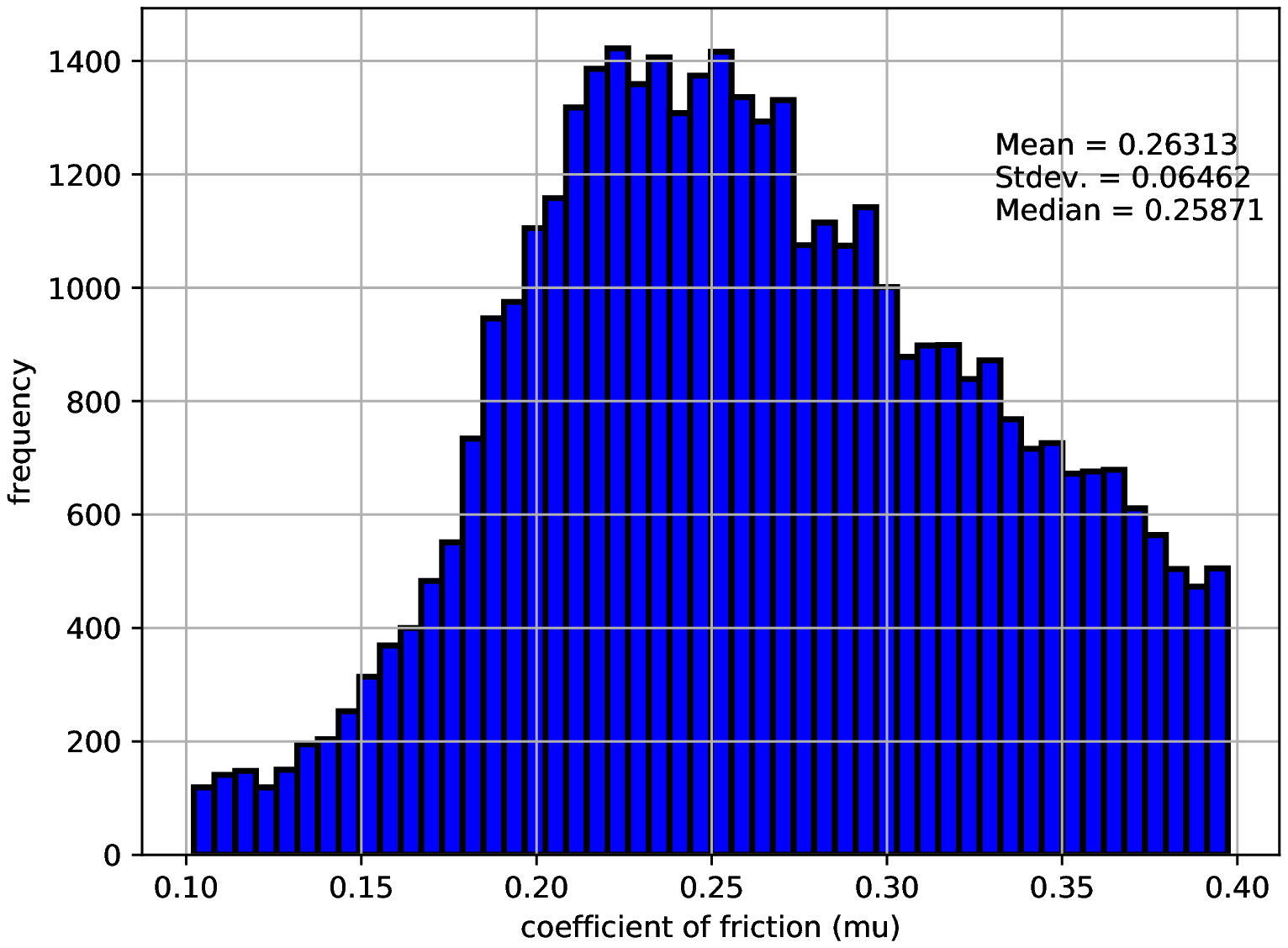}}
  \subfigure[]{\includegraphics[width=0.5\textwidth]{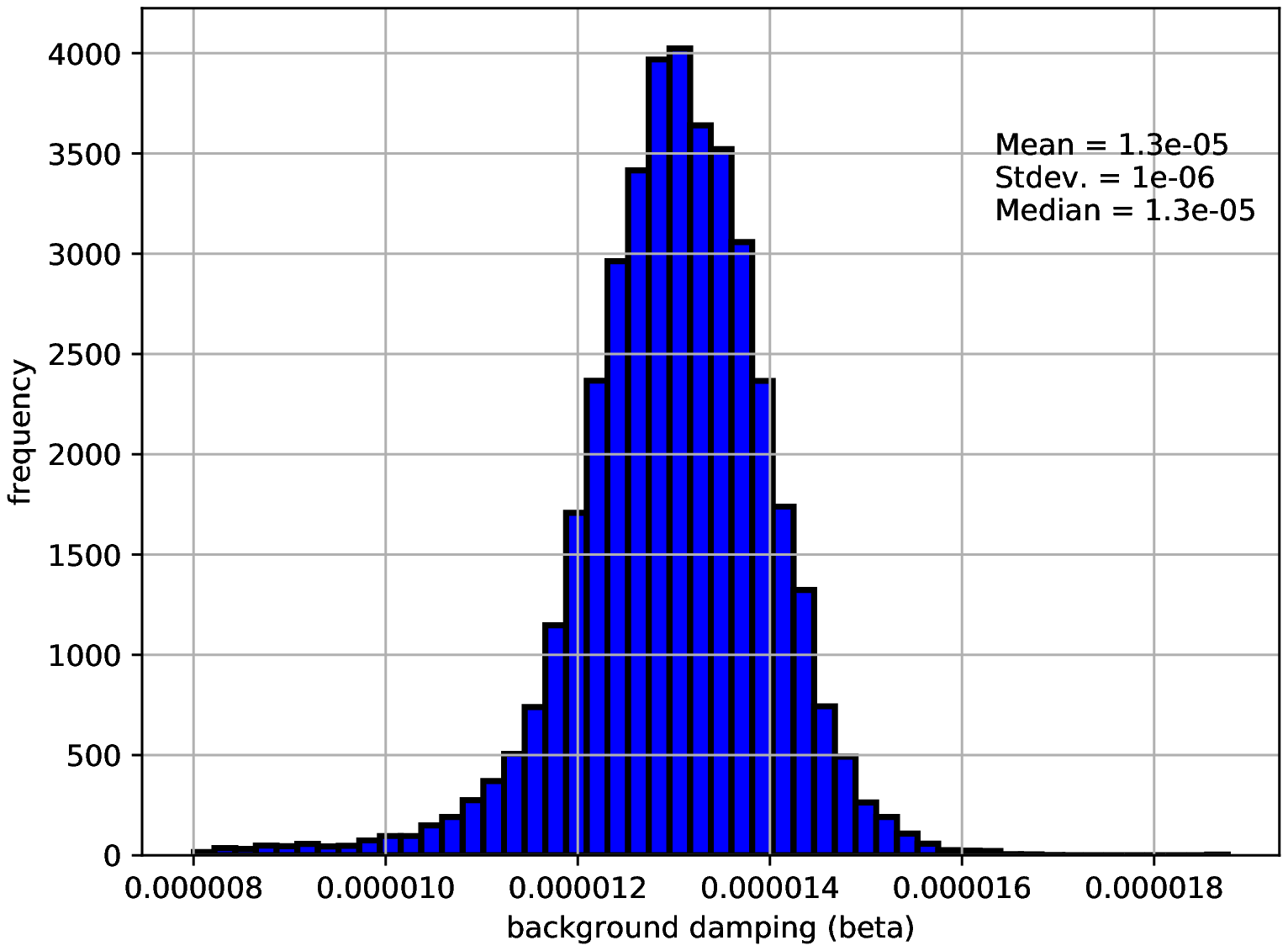}}
  \caption{Calibration results for the two parameters  Subfigure~(a) coefficient of friction, and Subfigure~(b) background damping.}
  \label{fig:calibration_vibration}
\end{figure}

\begin{figure}[h]
  \subfigure[]{\includegraphics[width=0.5\textwidth]{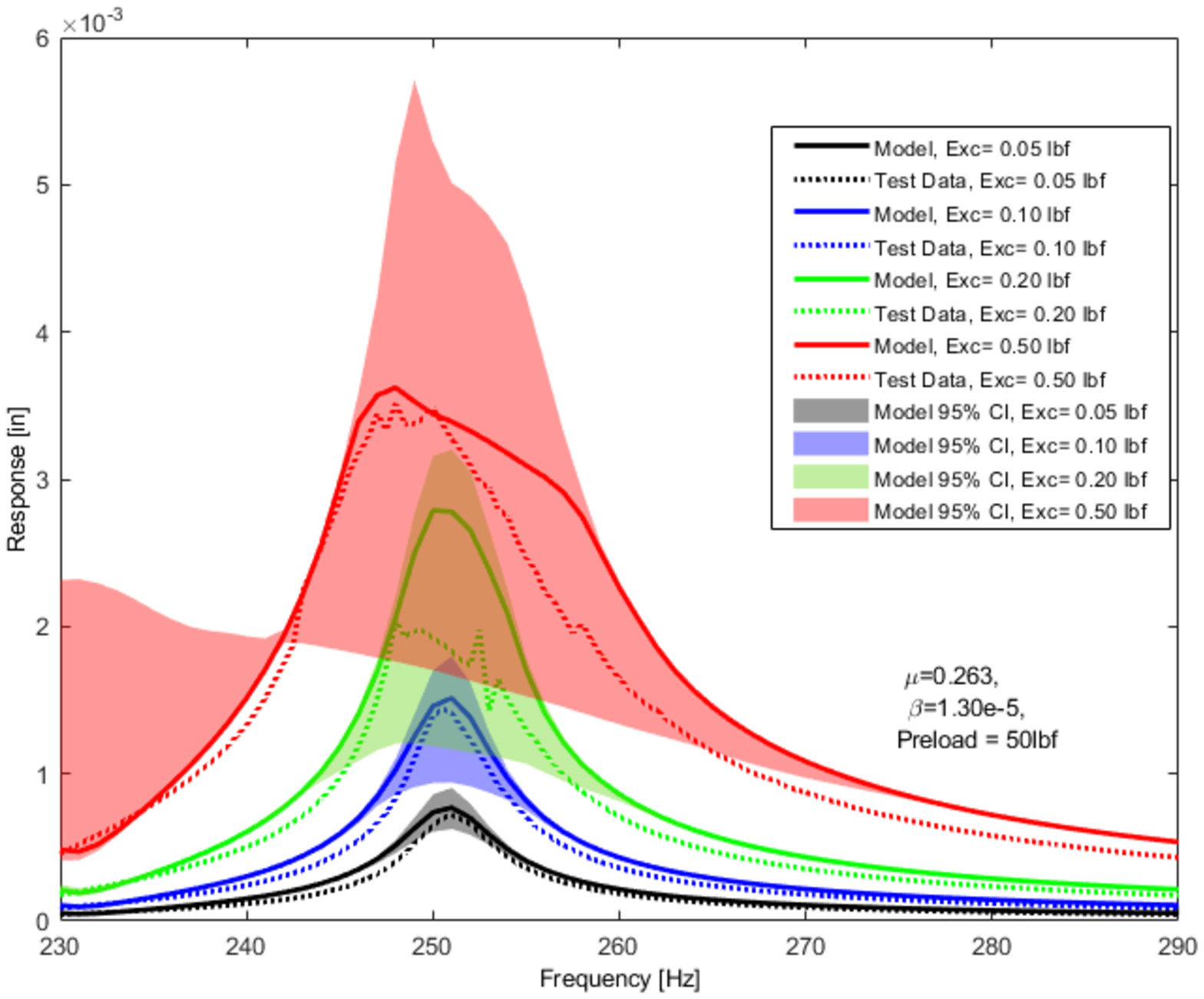}}
  \subfigure[]{\includegraphics[width=0.5\textwidth]{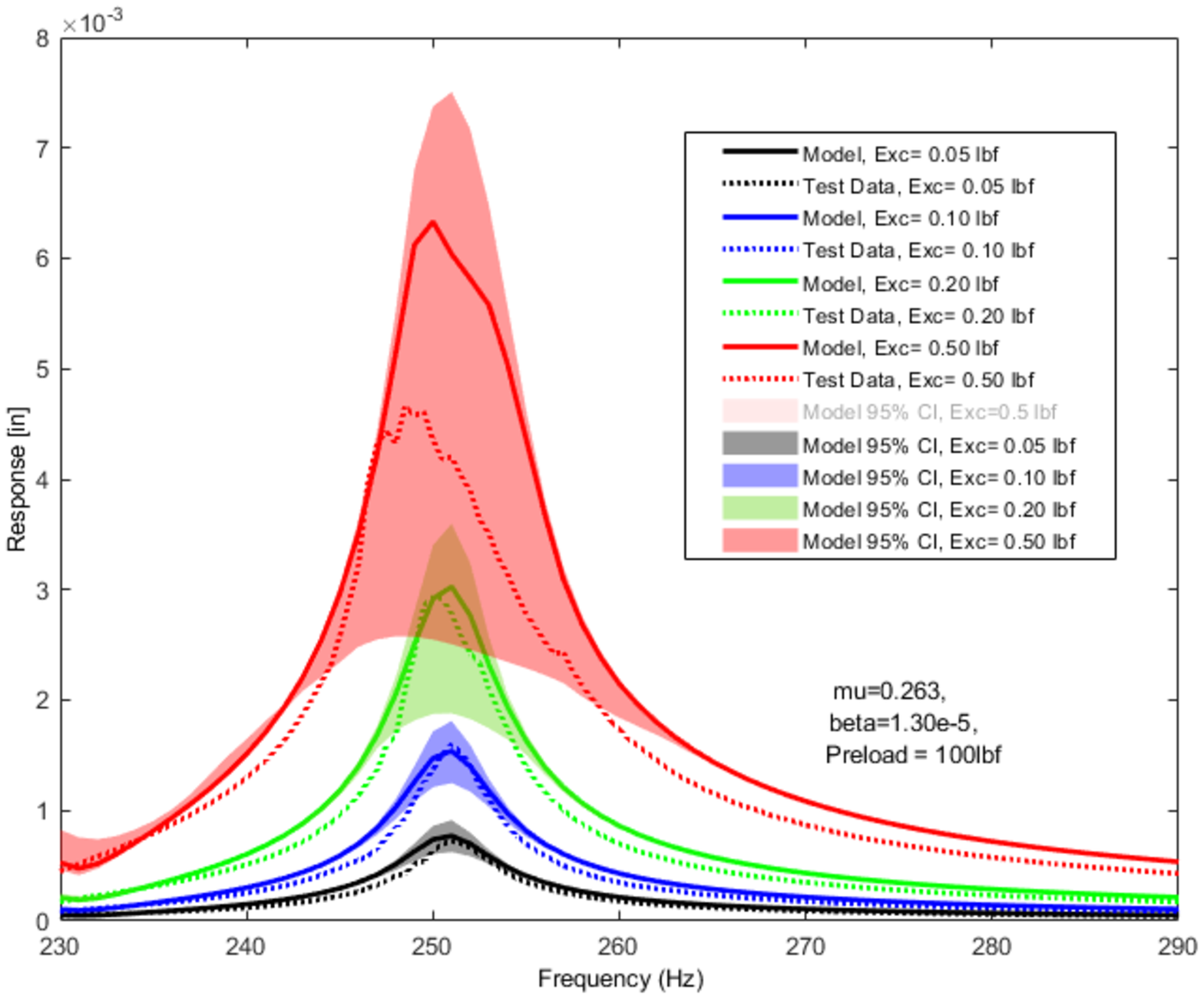}}
  \caption{Validation results for the two cases  Subfigure~(a) 50 lbf preload, and Subfigure~(b) 100 lbf preload.}
  \label{fig:calibration_validation}
\end{figure}

The nonlinear vibration analysis is run at the test conditions using the mean values of the calibrated parameters $\mu$ and $\beta$. Fig.~\ref{fig:calibration_validation} shows the comparison of the calibrated model results and the test data at two different preloads (50 lbf and 100 lbf). For the 50 lbf case, analytical responses obtained with the calibrated parameters match test data well in terms of overall frequency response for three out of four dynamic excitation levels (0.05, 0.1, 0.5 lbf). For the  0.2 lbf excitation the analytical response overpredicts the test data by 30$\%$. The frequency at which the maximum response occurs is predicted accurately for all excitation levels. For the the 100 lbf case, good matches on the maximum response are obtained for three out of four forcing levels. For the 0.5 lbf forcing level the model overpredicts the max response by 37$\%$. In all other cases, maximum response prediction errors are less than 5$\%$. It is possible to draw samples from the distribution of $\mu$ and $\beta$ and run the forced response analysis on these samples to generate confidence bounds on model predictions if the computational cost does not become prohibitive. This is illustrated in  Fig.~\ref{fig:calibration_validation}(b) where it can be seen that most of the test data falls within 95$\%$ model prediction bounds. Overall, with the calibrated parameters good match between data and model was observed over a wide range of preload and dynamic excitation levels.

\section{Advances and Applications}
\label{Applications}
\subsection{Transient Response}

In this section, we incorporate the ability to handle multiple transient responses in the GEBHM framework. We only present a brief introduction to the capability here and refer the interested reader to \cite{transientNatarajan2013} for more information on the theory and industrial applications. Similar to the original GEBHM framework, we utilized Higdon et al's \cite{Higdon2008} implementation of the transient calibration capability. Instead of dealing with the outputs directly in the original GEBHM implementation, we initially perform a singular value decomposition (SVD) of the outputs to identify the principal components to reduce the overall dimensionality of dealing with large transients variables, as follows:
\begin{align}
\label{eq:svd}
& \eta(\bm{X},\bm{\theta},t)_{n_t\times m} = U_{n_t\times n_t}\Sigma_{n_t\times m}V^T_{m\times m},
\end{align}

where $n_t$ are the number of  time-points in the transient outputs, $m$ are the number of simulations, and $\bm{\theta}$ are the calibration parameters of the simulation model $\eta$, $U$ and $V$ are the left and right singular vectors, and $\Sigma$ is the diagonal matrix of singular values. Then we retain only certain number of principal components (say $n_{p_u}$) to approximate the transient simulation model as,
 \begin{align}
\label{eq:eta_reddim}
& \eta(\bm{X},\bm{\theta},t)_{n_t\times m} \approx U_{n_t\times n_{p_u}}\Sigma_{n_{p_u}\times n_{p_u}}V^T_{n_{p_u}\times m},
\end{align}
which can then be further linearized as,
\begin{align}
\label{eq:eta_lin}
& \eta(\bm{X},\bm{\theta},t) \approx  K_{sim}w,
\end{align}
where $K_{sim}$ is $U\Sigma$ and $w$ is $K^{-1}_{sim}\eta$. Next we build a GP model for the principal components $w$ term. The simulation output, the experimental output can also be linearized as,
\begin{align}
\label{eq:y_lin}
& y(\bm{X},t) \approx  K_{obs}u,
\end{align}
where $K_{obs}$ is computed by interpolating $K_{sim}$ from the simulation time grid onto the observed time grid. This ensures generality of the overall method such that the simulation and observed data are measured at different points in time. The components of the observed data  is modeled as a GP with covariance matrix as shown in Eq.~\ref{eq:sigma_y}.

In order to model the discrepancy between the simulation model and observations for the transient outputs, we utilize a specific number of Gaussian basis for the linear model as,
\begin{align}
\label{eq:delta_lin}
& \delta(\bm{X},\bm{\theta},t) = \Delta(t) v(\bm{X}),
\end{align}

 
Discrepancy components $v$ are modeled again as a GP similar to that shown in Eq.~\ref{eq:sigma_delta}. Essentially, we have utilized linearization  to represent the transient problem using principal components for the simulation model, and Gaussian basis components for the discrepancy model in order to develop the GP models. Once these covariance functions are constructed, the hyperparameters can be inferred in a similar manner as described in the earlier sections. We have enhanced Los Alamos's original implementation to handle multiple transient outputs and have demonstrated good success in applying it to industrial applications, as shown in \cite{transientNatarajan2013}.

 \begin{figure}[h]
\begin{center}
 \includegraphics[width=0.5\textwidth]{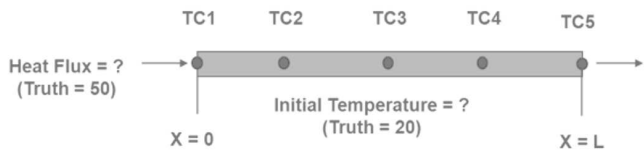}
 \end{center}
  \caption{Schematic of the transient thermal model showing the location of the observed data}
    \label{fig:thermal}
\end{figure}

\subsubsection*{Application: Transient Response}

\begin{figure}[h]
\subfigure[]{\includegraphics[width=0.5\textwidth]{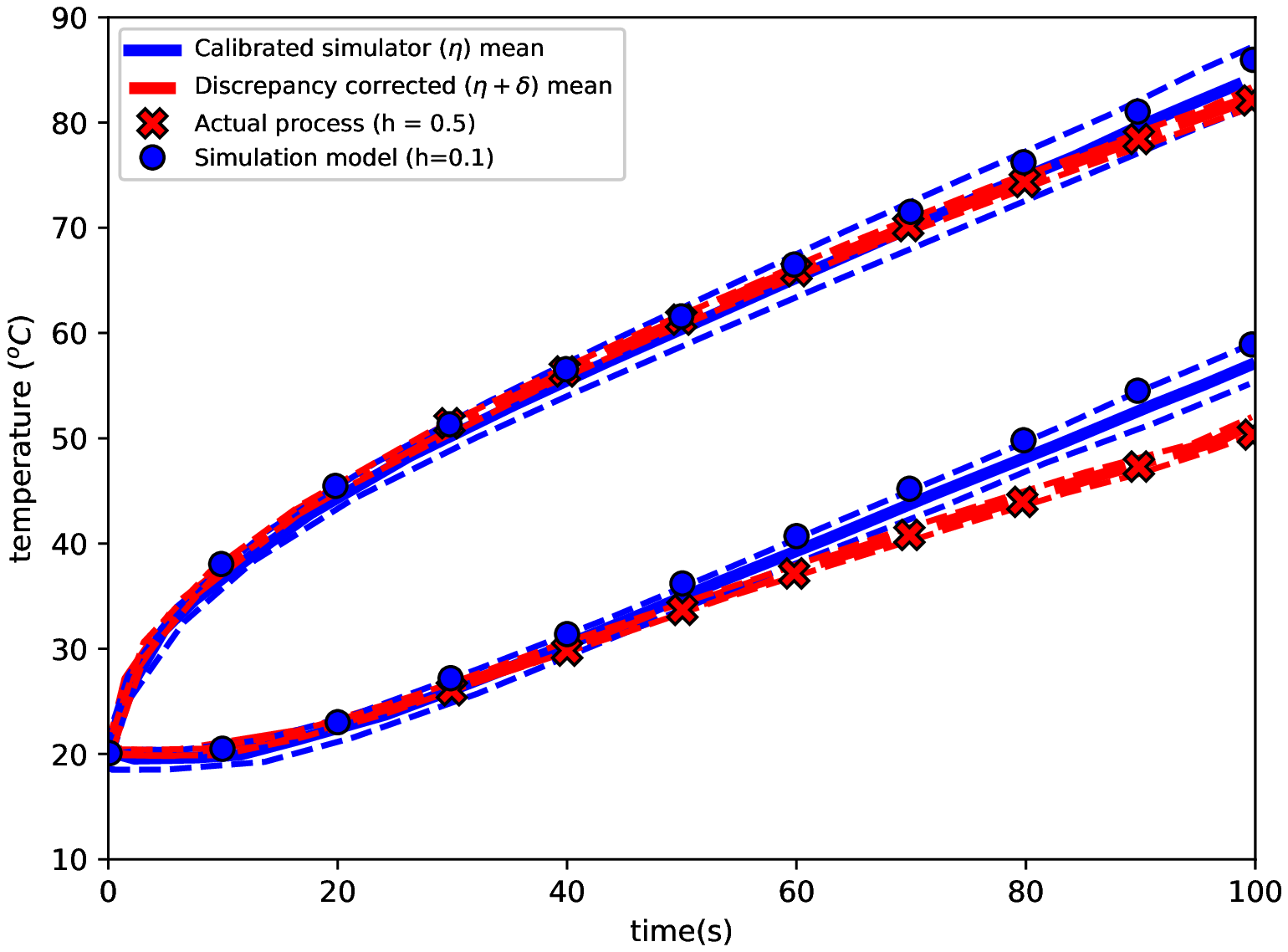}}
  \subfigure[]{\includegraphics[width=0.5\textwidth]{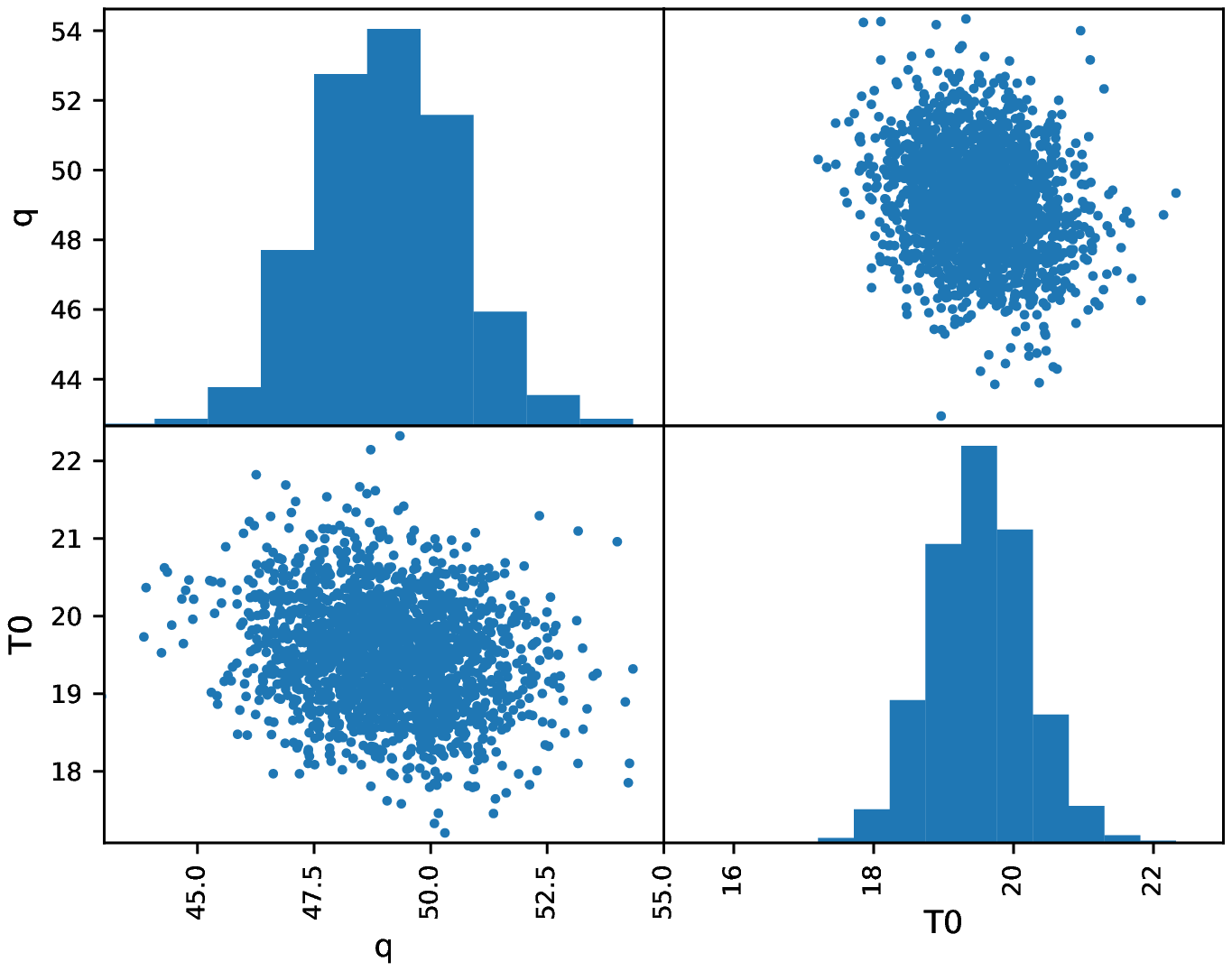}}
  \caption{(a) Histogram and scatter plot showing the posterior distributions of the two calibration parameters, and (b) Plot showing the calibrated simulator model and discrepancy model for two locations.(figures  reproduced from \cite{transientNatarajan2013})}
  \label{fig:response}
\end{figure}

We provide a brief description of one of the problems addressed in \cite{transientNatarajan2013}, which is a transient heat transfer problem on a 1D rod. Let us consider a rod of length $L$, as shown in Fig.~\ref{fig:thermal} with a heat flux on the left end and an initial temperature condition. Our aim here is to calibrate the heat flux and initial temperature based on the temperatures monitored on five different locations along the length rod. Fig. \ref{fig:response}(a) shows the calibrated posterior distributions of heat flux and the initial temperatures - we can clearly see that the GEBHM method can identify the strong correlation between the flux and initial temperature clearly. 
Figure \ref{fig:response}(b) shows the transient temperature signals with the discrepancy correction.  We have summarized the multiple transient calibration capability that is part of the overall GEBHM framework here, and interested readers are referred to our publication here: \cite{transientNatarajan2013}.
\subsection{Robust Gaussian Process}
In many industrial problem, the input $\bm{x}_*$ cannot be observed directly and the uncertainty around it is modeled with a Gaussian distribution as $\bm{x}_* \sim \mathcal{N}(\bm{u},\bm{S})$, with mean $\bm{u}$ and covariance $\bm{S}$.
In such a scenario is it important to quantify the effect of input uncertainty on the output, which is generally done by uncertainty propagation using a meta-model such as a GP.
The resulting predictive distribution of the the output, $p(y_*)$, when the input is a random variable $\bm{x}_*$ given by $p(\bm{x}_*|\bm{u},\bm{S})$, is obtained by marginalizing over the input distribution:
\begin{align}
\label{eq:preddist}
& p(y_*|\bm{u},\bm{S},\mathbb{D})= \int p(y_* | \bm{x}_*,\mathbb{D}) p(\bm{x}_*|\bm{u},\bm{S})d\bm{x}_* \; , \\
& \nonumber \\
& p(y_*| \bm{x}_*,\mathbb{D})= \mathcal{N}(\bar{y}_*,\sigma_{y_*}) \; ,
\end{align}
where $\mathbb{D}$ is the training data, $\bar{y}_*$ and $\sigma_{y_*}$ are the predicted mean and standard deviation of output. 

\begin{figure}[t]
	\begin{center}
		\includegraphics[scale=0.5]{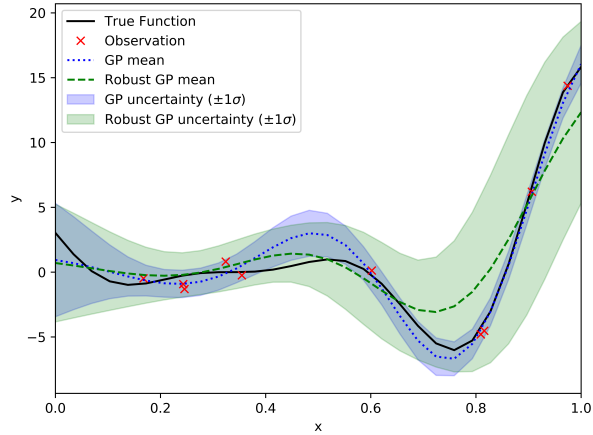}
	\end{center}
	\caption{Example on uncertainty prediction using standard GP and robust GP (Standard GP: $x\in[1,10]$ , Robust GP: $x\sim\mathcal{N}(u,0.01)$, $u\in[1,10]$). (figures  reproduced from \cite{ryan2018gaussian})}
	\label{fig:robGP1} 
\end{figure}

Traditionally, marginalization is estimated by Monte-Carlo sampling approach. The sampling approach using GP is fast, however, in a scenario like uncertainty-based design optimization, the {\it double-loop} process (optimization loop and sampling loop) can make the overall process expensive when the system has a large number of random input variables.
Robust GP using GEBHM overcomes this challenge by collapsing the double-loop uncertainty based design optimization problem into a single-loop problem. This is accomplished by exactly calculating the uncertainty measures $E_{\bm{x}}[y_*]$ and $V_{\bm{x}}[y_*]$. The robust GP can directly, with one function evaluation, estimate the first and second moments of the output with the consideration  the effect of input uncertainty around. An example of this distinction is shown in Fig.\ \ref{fig:robGP1}.  Note $\bar{y}_*$ and $\sigma_{y_*}$ are probabilistic GP model predictions and cover model uncertainty only, not the added uncertainty due to uncertain inputs.  

Assuming that $\bm{x}_*$ is a normally distributed random variable, GEBHM can be used to obtain exact analytical expressions for the mean, $m(\bm{u},\bm{S})$, and variance, $v(\bm{u},\bm{S})$, of the marginalized predictive distribution of output. 
This can enable the formation of single-level uncertainty-based optimization problems, where the objective is some function of $m(\bm{u},\bm{S})$ and/or $v(\bm{u},\bm{S})$.

The method to solve $m(\bm{u},\bm{S})$ and/or $v(\bm{u},\bm{S})$ implemented in GEBHM is based on work developed by Quinonero-Candela, \textit{et al.} \cite{Candela2003,Candela2003b,Girard2004}.
Using the law of iterated expectations, and the law of total variance:
\begin{align}
m(\bm{u},\bm{S}) &= E_{\bm{x}}[E_y[y|\bm{x}]]=E_{\bm{x}}[\bar{y}] \;, \\
& \nonumber \\
v(\bm{u},\bm{S}) &= E_{\bm{x}}[V_y[y|\bm{x}]] + V_{\bm{x}}[E_y[y|\bm{x}]] \; , \nonumber \\
&= E_{\bm{x}}[\sigma_{y}^2]+V_{\bm{x}}[\bar{y}] \; ,
\end{align}
where $E_{\bm{x}}[\cdot]$, $V_{\bm{x}}[\cdot]$ denote the expectation and variance wrt. $\bm{x}$, respectively. When using Gaussian kernels in GPs, the expressions for $\bar{y}$ are Gaussian  functions of $\bm{x}$ and the expressions for $\sigma_{y}^2$  are products of Gaussian  functions of $\bm{x}$.  Therefore, the integrand involved in determining $m(\bm{u},\bm{S})$ and $v(\bm{u},\bm{S})$ are products of Gaussian  functions, which allows for an analytical calculation.
The analytical form of the mean is given as: 
\begin{align}
\label{eq:mrobust}
 m(\bm{u},\bm{S}) &= \bm{\Psi}^\top \bm{l} \; , \\
 \nonumber
 \bm{\Psi} &= \left(\Sigma^{\eta}\right)^{-1}\bm{y} \;.
\end{align}
Elements of vector $\bm{l}=[l_1,\ldots,l_m]^\top$ are given by,
\begin{align}
\label{eq:lvec}
l_j = &\frac{1}{\lambda_{\eta_z}}\begin{vmatrix}2\left(\mathbf{B}\right)^{-1}\bm{S}+\bm{I}\end{vmatrix}^{-\frac{1}{2}} \nonumber\\
& \times\exp(-(\bm{u}-\bm{x}_j)^\top(2\bm{S}+\mathbf{B})^{-1}(\bm{u}-\bm{x}_j))+ \bm{I}\frac{1}{\lambda_{\eta_s}} \; ,
\end{align}
where $\mathbf{B} = diag[\beta_{y1}^{-1},\ldots,\beta_{yp}^{-1}]$ and $\bm{I}$ is a $m$ by $m$ identity matrix. The variance is given by,
\begin{align}
\label{eq:robvar}
v(\bm{u},\bm{S}) = &\Sigma_{**}^{\eta}-\mbox{Tr}((\left(\Sigma^{\eta}\right)^{-1}-\bm{\Psi}\bm{\Psi}^\top)\bm{L}) -\mbox{Tr}(\bm{l}\bm{l}^\top\bm{\Psi}\bm{\Psi}^\top) \; ,
\end{align}
\noindent
where, $\Sigma_{**}^{\eta}$ is the covariance matrix constructed using \cref{eq:sigma_eta} for new design point $\bm{x}_*$,  $\Sigma_{**}^{\eta} = \Sigma^{\eta} (\bm{x}_*,\bm{x}_*) $. $\mbox{Tr}(\cdot)$ is the trace operator and $\bm{L}$ is a $m$ by $m$ matrix with elements given by the following:
\begin{align}
L_{i,j}  &=\left(\frac{1}{\lambda_{\eta_z}}\right)^2\begin{vmatrix}4\left(\mathbf{B} \right)^{-1}\bm{S}+\bm{I}\end{vmatrix}^{-\frac{1}{2}} \nonumber \\
&\times\exp\left(-\left[(\bm{u}-\bm{x}_d)^\top\left(2\bm{S}+\frac{\mathbf{B}}{2}\right)^{-1}(\bm{u}-\bm{x}_d)\right.\right. \nonumber\\
&+\left.\left.(\bm{x}_i-\bm{x}_j)^\top\left(2\mathbf{B}\right)^{-1}(\bm{x}_i-\bm{x}_j) \vphantom{\frac{\bm{\Lambda}}{2}}\right]\right)+ \bm{I}\frac{1}{\lambda_{\eta_s}}, 
\end{align}
\noindent
where we define $\bm{x}_d = \frac{1}{2}(\bm{x}_i+\bm{x}_j)$.

\subsubsection*{Application: Robust Gaussian Process}
\begin{figure}[h]
	\begin{center}
		\includegraphics[scale=0.5]{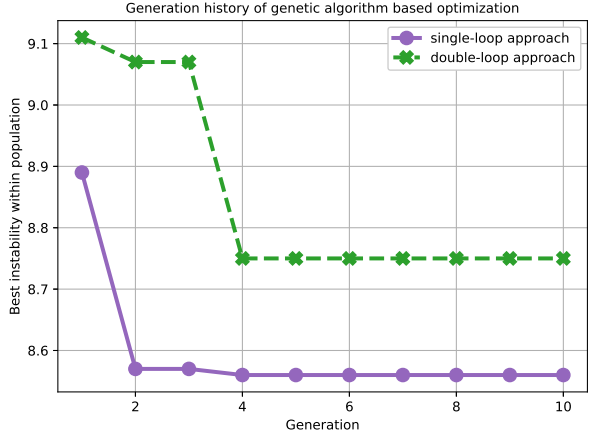}
	\end{center}
	\caption{Comparison of generation history of Genetic Algorithm for robust design optimization problem between double-loop approach and single-loop approach (figures  reproduced from \cite{ryan2018gaussian})}
		\label{robostGP}
\end{figure}

An application presented here is a customized subset of the proposed NASA UQ Challenge~\cite{Crespo2014}.
The optimization problem is to minimize the instability with a constraint on the magnitude of the instability uncertainty.
The instability is a function of four Gaussian random variables ($\bm{x} = [d_6,d_{13},p_4,p_5]$). The design variables are the respective mean values of ($d_6,d_{13},p_4,p_5$). For optimization, a Genetic Algorithm (GA) was used with a population of 100, and set to run for 10 generations.  At each optimization step, the single-loop approach calculates $E_{\bm{x}}[\mbox{Instability}]$ and $V_{\bm{x}}[\mbox{Instability}]$ directly from a robust GP with a single function evaluation. In comparison, with the double-loop approach, uncertainty metrics are calculated using $1000$ Monte-Carlo samples from a standard GP. The time-history of GA's iterations for the double-loop and single-loop approaches, are provided in Figure \ref{robostGP}.  The robust optimum was found between generations 3 and 4 for both approaches.  The single-loop approach takes 336 function calls, while the double-loop approach reaches the minimum instability value after 347,000 function calls. The  double-loop approach costs about 1000 times more than the single-loop approach  to find the robust optimum. For more details on theory and the application on robust GP, please refer to \cite{ryan2018gaussian,ghosh2020efficient}.


\subsection{Bayesian Sobol Sensitivity Analysis}
\label{sec:bsa_gebhm}
Design and analysis of complex engineering systems often presses for a 
thorough know-how of the relative order of importance of the 
various inputs. 
In this work, the model of an engineering system could be a physical experiment 
and/or a computer simulation which can be evaluated at a specific set of input conditions.
Such models are known as expensive black-box functions.
Additional challenges usually associated with such models are a high-dimensional input space
and a lack of gradient of information.
The inputs to the model can include some design variable(s) and some unknown parametric 
constant(s) of the model.
Knowledge of the relative sensitivity of the inputs is critical for designers
while making decisions under uncertainty, as it helps in expediting design exploration, 
identifying a set of critical variables and making simulation decisions under limited-data.


The variance-based methods, commonly known as the Bayesian Sobol sensitivity analysis (BSSA),
are more flexible to undertake the task of obtaining sensitivity information
for high dimensional problems with limited data under uncertainty.
The BSSA investigates how the uncertainty in the black-box model of a system can be divided among the different
uncertain inputs and their interactions.  An addendum to the above facet of the Sobol sensitivities is the ability 
to infer sensitivity of the underlying physical response to higher-order interactions.
All expressions for the main-order and higher-order effects are
computed using analytical expressions. 


We assume that the inputs are uncorrelated and have the following distributions.
\begin{equation}
\label{eqn:input_dist}
x_{i} \sim \mathcal{U}(0,1) \forall i \in d.
\end{equation}
where $d$ is the dimensionality of the input space and $ \mathcal{U}(0,1)$ is a standard uniform distribution. Since we do not have full access to the true response, we resort to using a probabilistic 
representation of $y(\bx)$ as described in \qref{koh_y}. Denoting the posterior predictive mean of the predictive simulator by $\eta(\bx; \btheta)$, where  $\bx$ represents the  design variable(s) and $\btheta$ denotes the calibration parameter(s), we make use of the high-dimensional model representation (HDMR)~\cite{sobol2001global} to represent $\overline{y}(\bx)$ as follows:

\begin{align}
\label{eqn:hdmr}
y(\bx) &=\eta(\bx; \btheta)\\
\nonumber
&=z_{0} + \sum_{p=1}^{d}z_{p}(x_{p}) + \sum_{p<q}z_{p, q}(x_{p}, x_{q})\\
\nonumber
& + z_{p,q,\cdots,d}(x_{p}, x_{q},\cdots,x_{d}),
\end{align}

where,
\begin{align}
\label{eqn:effect_funcs}
&z_{0} = \mathbb{E}(Y),  \\
&z_{p} = \mathbb{E}(Y|x_{p}) - \mathbb{E}(Y), \nonumber \\
&z_{p, q} = \mathbb{E}(Y|x_{p, q}) - z_{p} - z_{q} -  \mathbb{E}(Y), \nonumber \\
&z_{p, q, r} = \mathbb{E}(Y|x_{p, q, r}) - z_{p, q} - z_{q, r} - z_{r, p} - z_{p} - z_{q} - z_{r} \nonumber
-\mathbb{E}(Y).
\end{align}

To avoid confusion between design variables and calibration parameters we concatenate $\bx$ and $\btheta$ to represent the set of variables as $\bx =\{\bx, \btheta\}$.
This is consistent with the formulation since the BSSA follows the same analytic formulation for both sets of variables.
The  terms denoted by $z$ in \qref{hdmr} are known as effect functions and are orthogonal under the assumption that the distribution over the input space is uniform and that the true response is square-integrable~\cite{sobol2001global}.
This  property simplifies the derivation and computation of the BSSA indices for the scenario with uncorrelated inputs.
\begin{equation}
\label{eqn:main_eff_expect}
\mathbb{E}(Y|\bx_{p}) = \int_{\mathscr{X}_{-p}}\eta(\bx)\du_{-p|p}(\bx_{-p}|\bx_{p}), 
\end{equation}

\begin{equation}
\label{eqn:def_sobol_p}
S_{p} = \frac{var\{\mathbb{E}(Y|\bX_{p})\}}{var\{Y\}}, 
\end{equation} 

\begin{equation}
\label{eqn:def_sobol_pq}
S_{pq} = \frac{var\{\mathbb{E}(Y|\bX_{pq})\}}{var\{Y\}}  - S_{p} - S_{q}.
\end{equation} 

Analytical tractability of the Gaussian form of $\eta(\bx)$, derived in~\cite{oakley2004probabilistic,srivastava2017analytical}, allows one to further represent the BSSA indices as follows:

\begin{equation}
\label{eqn:def_sobol_gauss_p}
S_{p} = \frac{\mathbb{E}^{*}var\{\mathbb{E}(Y|\bX_{p})\}}{\mathbb{E}^{*}var\{Y\}} 
\end{equation} 
and two-way interactions as:
\begin{equation}
\label{eqn:def_sobol_gauss_pq}
S_{pq} = \frac{\mathbb{E}^{*}var\{\mathbb{E}(Y|\bX_{pq})\}}{\mathbb{E}^{*}var\{Y\}}  - S_{p} - S_{q}
\end{equation} 
\noindent where, $\mathbb{E}^{*}$ is the expectation with respect to the probabilistic surrogate model.

Similarly, we can define the BSSA for higher-order interactions as:
\begin{equation}
\label{eqn:def_sobol_gauss_pqr}
S_{pqr} = \frac{\mathbb{E}^{*}var\{\mathbb{E}(Y|\bX_{pqr})\}}{\mathbb{E}^{*}var\{Y\}}  - S_{pq} - S_{qr}
- S_{rp} - S_{q} - S_{p} - S_{r}
\end{equation}

\subsubsection*{Application: Bayesian Sobol Sensitivity Analysis}
\label{sec:bsa_gebhm_app}
The structural dynamics problem discussed in the previous section is used as a test-bed to demonstrate the BSSA capability of GEBHM. The BSSA obtained with respect to the design variables and the calibration parameters are shown in \fref{bssa_vibration}.

\begin{figure}[h]
	\subfigure[]{\includegraphics[width=0.5\textwidth]{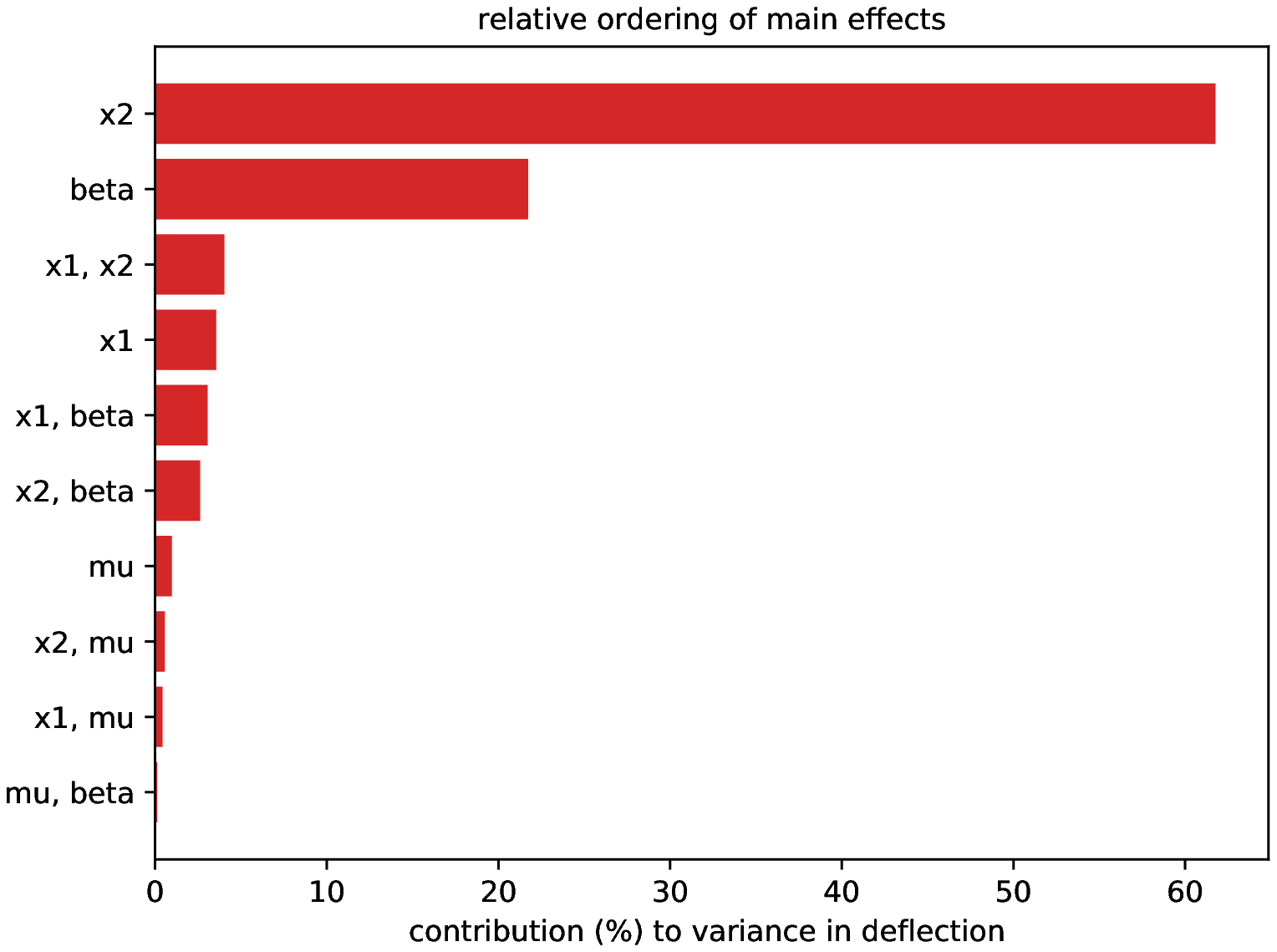}}
	\subfigure[]{\includegraphics[width=0.5\textwidth]{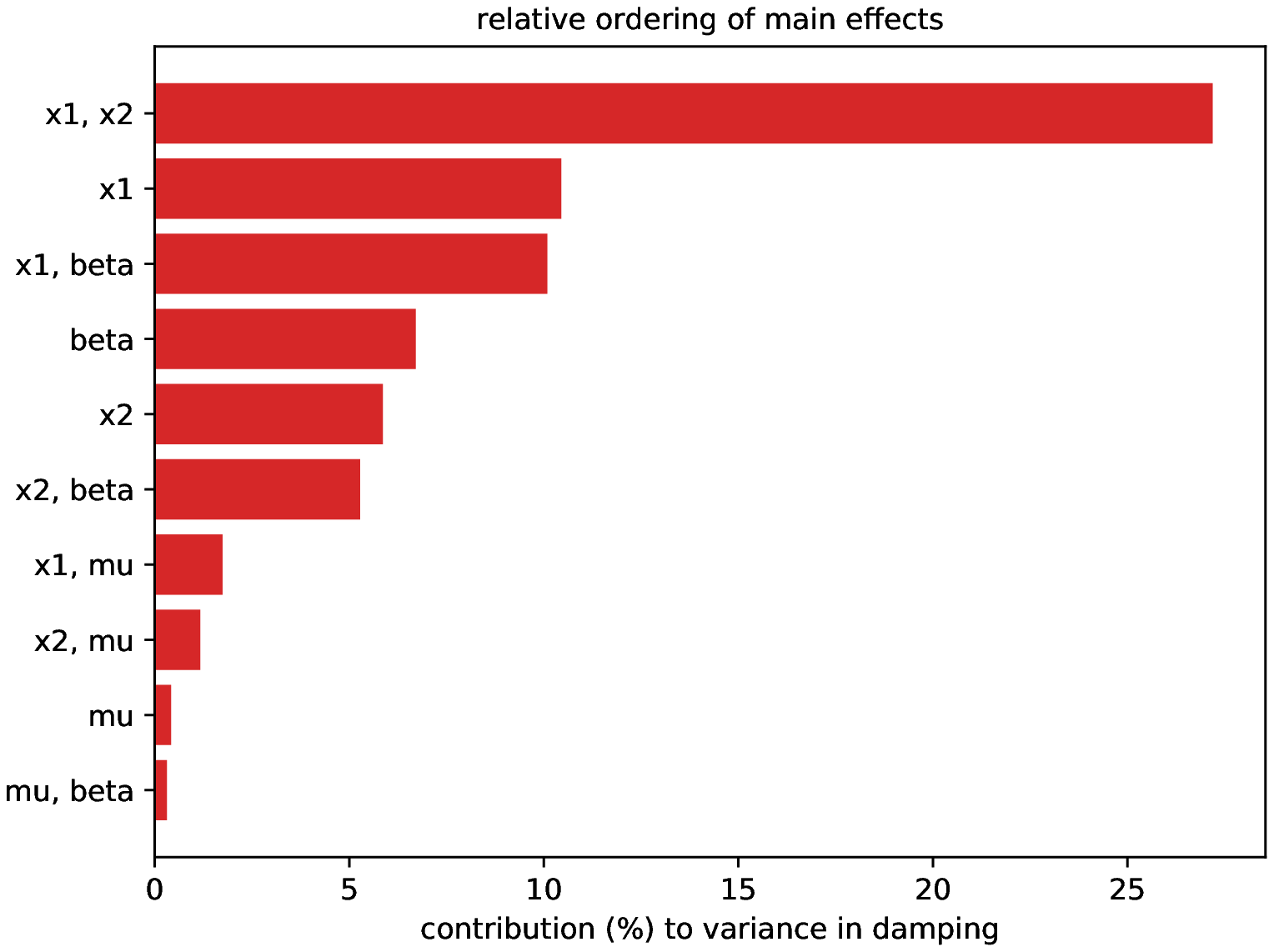}}
	\caption{The Bayesian Sobol sensitivity analysis (BSSA) for the (a) deflection and (b) damping in the vibration problem.}
		\label{fig:bssa_vibration}
\end{figure}

\subsection{Bayesian Sobol Sensitivity Analysis with Correlated Inputs}
\label{sec:bsa_gebhm_corr}
The work discussed in the previous sections assumes that inputs are uncorrelated.
In that case, the HDMR has orthogonal effect functions, and
one may compute each Sobol index independently.
Presence of correlated inputs mandates sampling from joint distributions of the inputs which increases the computational cost tremendously.
At the same time,  ignoring correlations can have a detrimental effect on the inferred BSSA indices as has been highlighted in \cite{saltelli2002relative}.

Saltelli et. al. \cite{saltelli2002relative} and other authors
have described this effect to be carried over due to correlation
and stress the need to be careful when making conclusions
with Sobol indices when inputs are correlated.
The BSSA approach developed in Xu et. al. \cite{xu2008uncertainty} splits the 
contribution of an individual input
to the uncertainty of the model output into two components: correlated and uncorrelated.
This approach was carried forward by 
Li et al. \cite{li2010global} with a
relaxed HDMR concept to compute variance contribution
due to correlation and problem structure for non-linear models.
Chastaing et al. \cite{chastaing2015generalized} develop a hybrid method for decomposing the variance by utilizing the  hierarchical
orthogonality of component functions and the
Gram–Schmidt method.
A hybrid method using polynomial chaos expansions and copula method was proposed in \cite{sudret2008global}.

We proceed with the framework proposed by Li and Rabitz \cite{li2010global,li2012general} which enforces orthogonality only on the lower-order component functions, a technique commonly known as hierarchical orthogonality.
In other words, $z_{pq}(x_{p}, x_{q})$ is only required to
be orthogonal to $z_{p}(x_{p})$ and $z_{q}(x_{q})$ and not to any other component functions. 
Following the above approach, the output variance can be decomposed as follows:
\begin{gather*}
\label{eqn:effect_funcs_corr}
var\{Y\} = \sum_{\bp=1}^{2^{d}-1}var\{z_{\bp}\} + cov\{z_{\bp}, \sum_{\bq=1}^{2^{d}-1} z_{\bq} \}
\end{gather*}

Here, $\bp$ and $\bq$ are multi-indices that denote the different interaction terms.
The total index $S_{\bp}$ is only a sum of the
structural and correlative contribution.

\begin{align}
\label{eqn:effect_funcs_corr_strct}
S_{\bp}^{a} &= \frac{var\{z_{\bp}\}}{var\{Y\}},\\
\label{eqn:effect_funcs_corr_corlt}
S_{\bp}^{b} &= \frac{cov\{z_{\bp}, \sum_{\bq=1}^{2^{d}-1} z_{\bq}\}}{var\{Y\}},
\end{align} 
\noindent
where $S_{\bp} = S_{\bp}^{a} + S_{\bp}^{b}$. Note that, the structural and correlative component functions cannot
be estimated independently of each other in the presence of input correlation and more advanced methods would be needed to infer the BSSA indices for additive models.

The use of GEBHM's BSSA with uncorrelated and correlated inputs has been successfully demonstrated on several industrial applications\cite{srivastava2014hybrid,srivastava2015variance,kumar2013calibrating}.
The usefulness of this framework is best highlighted in problems with reasonably high-dimensionality and limited observed or simulation data under uncertainty, while being fully integrated with GEBHM's model calibration framework.
We demonstrate the approach on industrial applications in the following sections.

\subsubsection*{Application: Bayesian Sobol Sensitivity Analysis with Correlated Inputs}
\label{sec:bsa_gebhm_corr_app}

To highlight the impact of the correlated BSSA capability of GEBHM we apply the framework on a problem of engine vibration.
The problem has one output and six design variables.
The correlation exists between inputs denoted by $x_1$ and $x_4$. 
The impact of including the correlative term on the overall ordering of the BSSA is highlighted in \fref{bssa_vibration_corr} (a) and (b).
The  structural BSSA is unable to incorporate the effect of the correlation between the two.
However, the final BSSA, with the correlative part included, reorders some of the interaction effects involving $x_1$ and $x_4$.

 \begin{figure}[h]
  \subfigure[]{\includegraphics[width=0.5\textwidth]{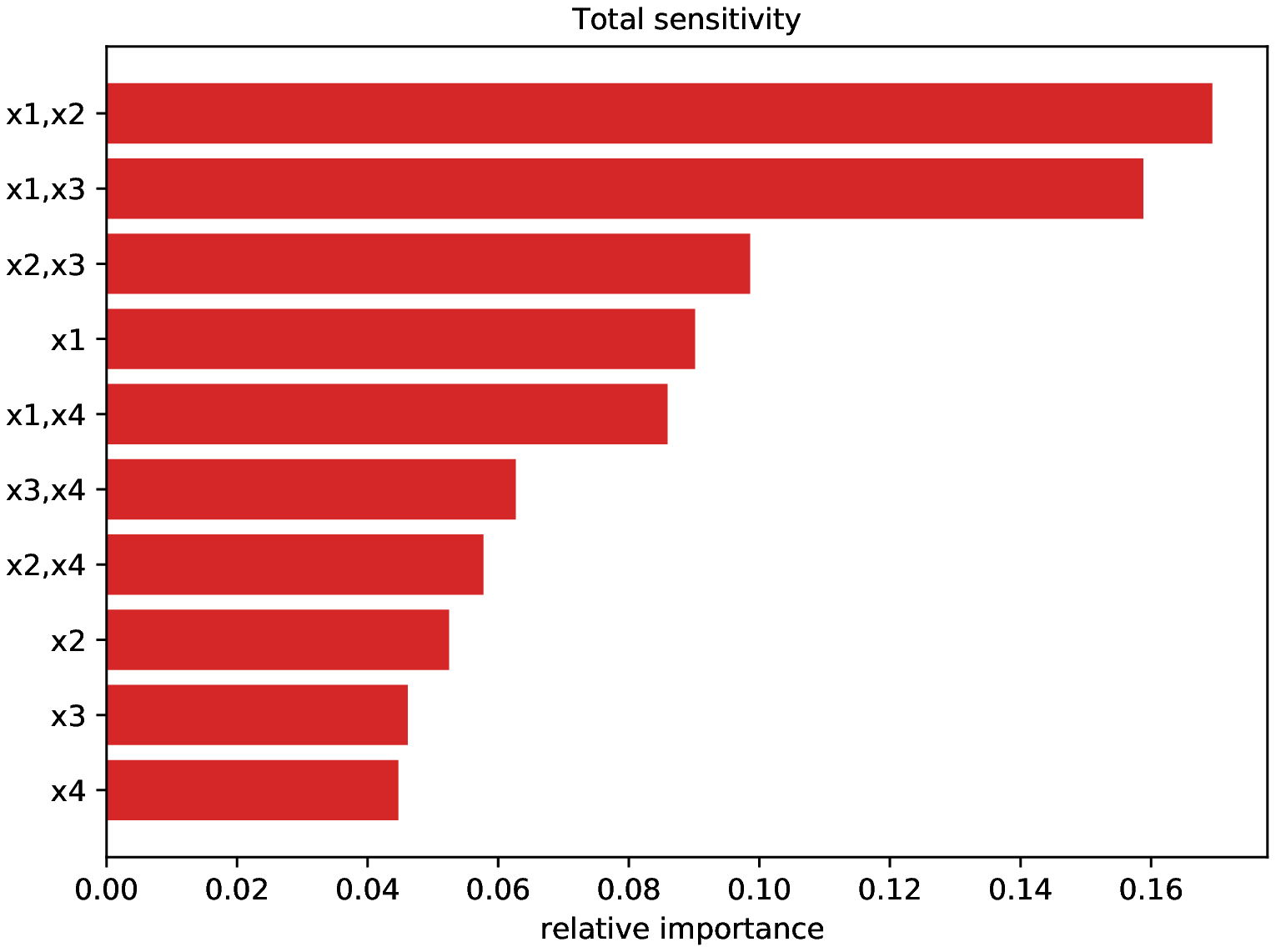}}
  \subfigure[]{\includegraphics[width=0.5\textwidth]{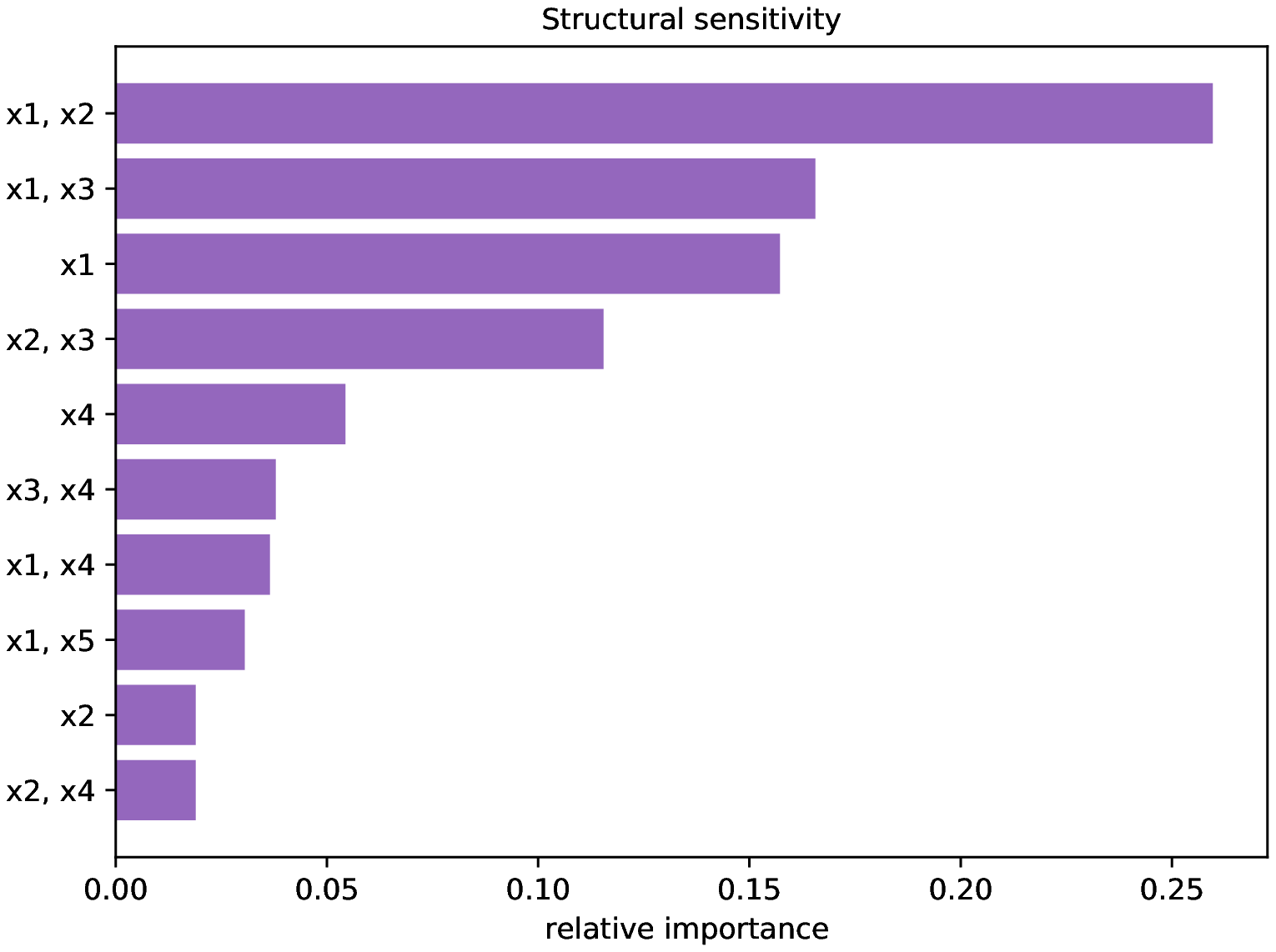}}
  \caption{The total (a) and the structural (b)  Bayesian Sobol sensitivity analysis (BSSA) of the vibration problem}
  \label{fig:bssa_vibration_corr}
 \end{figure}

\subsection{Portable Gaussian Process}
\label{sec:portable_bhm}
It is often the case that a small number of effects as identified in Sec.\ \ref{sec:bsa_gebhm} explain the majority of the GEBHM metamodel's predictive trends.
In addition, we have encountered a number of applications where the $\order{m^2}$ storage footprint and $\order{m}$ computational cost associated with computing predictions with the full GEBHM make it infeasible to apply.
In this section, we turn our attention to a method for approximating the posterior mean $m$ of the GEBHM model by leveraging the observation that it may be approximated by the combination of a relatively small number of effect functions;
each of these effect functions is approximated through a generalized linear model.
Doing so eliminates the undesirable nonparameteric characteristics of the GEBHM while incurring a negligible penalty in predictive performance.
Due to the lightweight nature of this model, we refer to it as a  \emph{portable Gaussian process}  or  \emph {portable Bayesian Hybrid Model}  (pBHM)\cite{kristensen2018polynomial}.
The remainder of this section describes the methodology by which one constructs a pBHM.

The approach takes as input a trained GP model from GEBHM and the set $\Omega_x \subset \R^p$ over which one expects to make predictions, equipped with a density $p(\bx)$.
We consider a scalar posterior mean $\eta(\cdot)$ of the GP model, recognizing that one may create a multi-output model by repeating the process outlined in this section for the other $N$ output dimensions.
The output of the approach is a parametric approximation of $\eta(\cdot)$ that may be used within $\Omega_x$.

Carrying out the sensitivity analysis of Sec.\ \ref{sec:bsa_gebhm}, we compute the Sobol indices of the HDMR of Eq.\ (\ref{eqn:hdmr}) and identify those whose magnitude are greatest; let the set of Sobol indices be represented by $\calS = \{ S_p \}$, where $S_p \subset \{1, \dots, p\}$ is the set of dimensions associated with an index.
We retain the effect functions corresponding to the largest Sobol indices, picking the number of terms such that the sum of the associated indices exceeds some threshold (typically 99\%), ensuring that this truncated series captures the majority of the variation in $\eta$ over $p(\bx)$.
Note that if the GP kernel function is of the form in Eqn.~\ref{eq:sigma_eta} and $p(\bx)$ factorizes into uniform or Gaussian marginals in each of the input dimensions, then the Sobol indices may be computed analytically.

As shown in Sec.\ \ref{sec:bsa_gebhm}, if $k(\cdot, \cdot)$ is parametric with an exponentiated quadratic form [e.g.\ as in Eq.\ (\ref{eq:sigma_delta})] and $p(\bx)$ is either uniform or Gaussian, then the effect functions may be computed analytically.
We now consider the task of fitting a parametric model to a given effect function $z_{S_p}$.
The following task is conducted independently for all $S_p$ contained in $\calS$ and may therefore be trivially parallelized.
We sample $\tilde m$ points from some experimental design over the input space in the dimensions contained in $S_p$.
This may come from a space-filling design (e.g.\ Latin Hypercube) or simply by sampling from the marginal $p(\bx_{S_p})$.
We collect the inputs 
$\tilde{\bX} \in \R^{\tilde m \times |S_p|}$ 
and corresponding outputs under the effect function 
$\tilde \by \in \R^{\tilde m}$ 
as a training dataset 
$\tilde \calD = \{\tilde{\bx}^{(i)}, \tilde y^{(i)} \}_{i=1}^{\tilde m}$.
We seek to learn a generalized linear model 
\begin{equation}
	z_{S_p} \approx \balpha_{S_p} \bphi_{S_p}(\bx_{S_p}),
	\label{eqn:pbhm:effect-function-approximation}
\end{equation}
where $\bphi_{S_p}$ contains a set of basis functions and $\alpha_{S_p}$ the corresponding coefficients.
We have used polynomial basis functions as well as splines in previous work \cite{kristensen2018polynomial}.
We determine $\balpha_{S_p}$ by minimizing a mean squared error loss on $\tilde \calD$ with an additional $L_1$ penalty to induce sparsity in $\balpha_{S_p}$;
the magnitude of the regularization hyperparameter is tuned by cross-validation. 
Note again that it is comparably cheap to generate data from the effect function to be approximated, so cross-validation is not subject to troubles arising from data scarcity.

%
%
%

\subsubsection*{Application: Portable Gaussian Process}
Here, we demonstrate the application of portable Gaussian process or pBHM to model crack growth for an airframe component.
We focus on constructing a surrogate model for a stress intensity factor (SIF), a key quantity driving crack growth, as a function of load and geometry. 
Because the surrogate will be used in a many-query setting, quick predictions are highly desired.

\begin{figure}[hbt]
 	\centering
 	\includegraphics[width=0.5\textwidth]{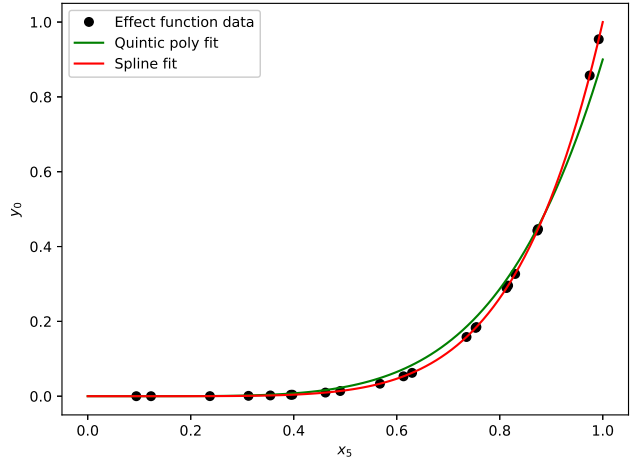}
	\caption{pBHM SIF modeling: Comparison of the effect function $z_5(x_5)$ for the SIF predicted by the source BHM model and several candidate parametric approximations. From \cite{kristensen2018polynomial}.}
	\label{fig:pBHM:effect-function}
\end{figure}

A GP model was trained using GEBHM on $m=360$ data generated via finite element simulations, parameterized by $p=7$ inputs.
A sensitivity analysis using GEBHM reveals that third- and higher-order interactions are negligible.
We consider several parametric fits for the effect functions, eventually choosing spline fits.
Fig.~\ref{fig:pBHM:effect-function} shows the fit obtained for a representative main effect function.
 
\begin{figure}[hbt]
	\centering
	\includegraphics[width=0.5\textwidth]{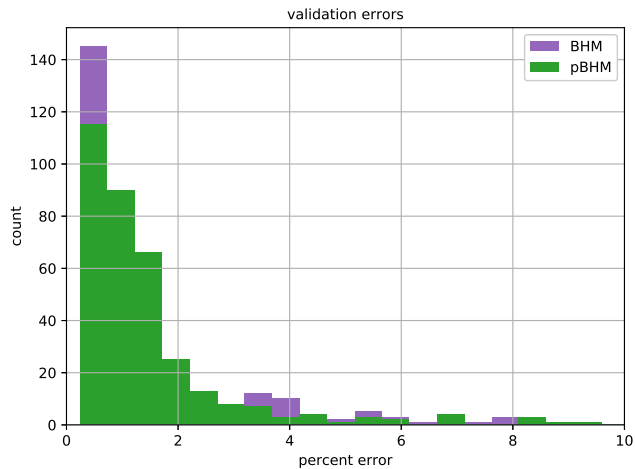}
	\caption{pBHM crack growth modeling: Comparison of the effect function $z_5(x_5)$ for the stress intensity factory predicted by the source BHM model and several candidate parametric approximations used by the pBHM model. From \cite{kristensen2018polynomial}.}
	\label{fig:pBHM:validation}
\end{figure}

The resulting spline model is tested on 360 held-out validation data.  
The percent errors of the original GP model from GEBHM model and the spline version are shown in Fig.\ \ref{fig:pBHM:validation}.  
The mean absolute error is 1.9\% for the GP and 2.9\% for the spline model. 
The root-mean-square-error is 5.2\% for the GP and 11.9\% for the spline model. 
In exchange for a modest decrease in accuracy, the the spline model enjoys a 50x speedup over the GP model.

\subsection{Multi-Source Legacy Data Modeling}

For a new industrial design, typically only sparse data is available due to cost and time associated with high-fidelity analysis and experimentations. For such complex systems, sparse data may not be sufficient to build an accurate model of design performance. Typically, a multi-fidelity approach is  sought when one can generate relatively-abundant data from a cheaper lower-fidelity model and integrate it with the sparse high-fidelity data. However, in an industrial setting, high-fidelity data are generally available from legacy designs which are not exactly same as the new design, but they belong to a similar family. In this case, we extend the capability to handle data from multiple legacy sources to improve the predictive capability of the new system. This enhancement can be applied even if low-fidelity data are available from legacy systems.

In multi-source legacy modeling with GEBHM, separate models are built for each legacy data by combining the legacy data with new system data.
For $k^{th}$ legacy data, the legacy model is built as $\mathbb{M}_k(\bm{x}) = {\eta}_k^{legacy}(\bm{x}) + {\delta}_k^{legacy}(\bm{x})$, by treating the data in $k^{th}$ legacy data source as simulation data and new system data as experimental or high-fidelity data.
It is assumed that the input variables (design, operational, etc.) and the output variables (performance, cost, etc.)  are the same for the new system and all the legacy systems.
In this scenario, when legacy-data has more variables than the new system, i.e. $dim(\bm{x}_{legacy}) > dim(\bm{x}_{new})$, then the extra variables in legacy data can be treated as calibration parameters.

If $K$ legacy models are generated, then the multi-source prediction at new location is evaluated as 
\begin{equation}
{y}(\bm{x}) = \sum_{k=1}^{K} {w}_k(\bm{x}) \left({\eta}_k^{legacy}(\bm{x}) + {\delta}_k^{legacy}(\bm{x})\right)
\end{equation}
\noindent where ${w}_k$ is the model validity which is a function of the input variables such that $\sum_{k=1}^{K}  {w}_k(\bm{x})=1$ for any value of $\bm{x}$.

The model validity associated with a legacy model signifies relative confidence of legacy models at a given design location $\bm{x}$. There can be multiple ways to evaluate model validity like expert opinion, Bayes' factor, relative age of the legacy data, etc. 
In GEBHM, we use two statistical metrics to evaluate model validity; likelihood-based model validity and  uncertainty-based model validity.
In the likelihood-based model validity, the validity of a legacy model at a input setting $\bm{x}_{new}^*$ where data for the new design is available is proportional to the probability of the legacy model to predict the output or performance of the new design.
For example, the likelihood of the $k^{th}$ legacy model, $\mathbb{M}_k$ at a location $\bm{x}_{new}^*$, for which output of new design is known to be $y_{new}$ is given as
\begin{equation}
w_{k}^{lik}(\bm{x}_{new}^*) \propto \frac{1}{\sqrt{2 \pi \sigma_k^2(\bm{x}_{new}^*)}}\exp \left(- {\frac{\left( y^* - \mu_k(\bm{x}_{new}^*)\right)^2}{2 \sigma_k^2(\bm{x}_{new}^*)}} \right)
\end{equation}
\noindent where $\mu_k(\bm{x}_{new}^*)$ and $\sigma_k(\bm{x}_{new}^*)$ are the predictive mean and standard deviation of the $k^{th}$ legacy model $\mathbb{M}_k$ at $\bm{x}_{new}^*$.
In the uncertainty-based model validity, the model validity at $\bm{x}_{new}^*$ is inversely proportional to the predictive uncertainty as
\begin{equation}
w_{k}^{unc}(\bm{x}_{new}^*) \propto \frac{1}{\sigma_k^2(\bm{x}_{new}^*)}
\end{equation}

The overall model validity of the $k^{th}$ legacy model at a given design point $\sigma_k^2(\bm{x}_{new}^*)$ is then given as
\begin{equation}
w_{k}(\bm{x}_{new}^*) = \kappa(\bm{x}_{new}^*) w_{k}^{lik}(\bm{x}_{new}^*) w_{k}^{unc}(\bm{x}_{new}^*).
\end{equation}
\noindent where $\kappa(\bm{x}_{new}^*)$ is a proportionality factor which is constant across all legacy model at a given $\bm{x}_{new}^*$.
GP models of model validity,  $w_{k}(\bm{x})$,  are then built for each legacy model by using the overall model validity evaluated at design points $\bm{x}_{new}$ at which new system data are available. 
For more details, please refer to \cite{ghosh2018bayesian}.

\subsubsection*{Application: Multi-Source Legacy Data Modeling}
\begin{figure}[hbt]
	\centering
	\includegraphics[width=0.5\textwidth]{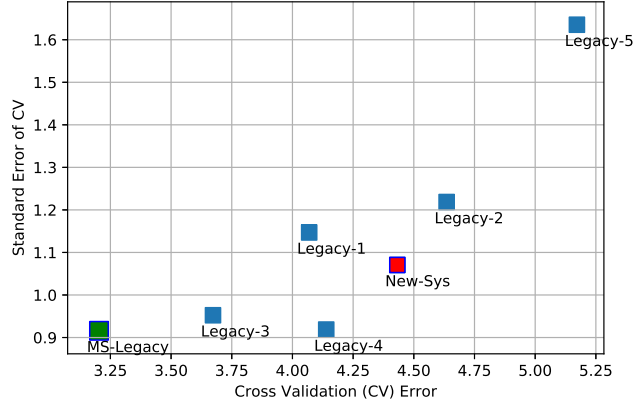}
	\caption{Cross validation error vs standard error for legacy data model compared with other models for pump maximum efficiency }
	\label{fig:pump_legacy}
\end{figure}

In one of the application, we have used multi-source legacy data modeling to predict pump efficiency for a new pump design.
These pumps are required for heavy duty services in refinery, petro-chemical and other industrial services, pumping fluids under wide range of pressures and particularly at high temperatures.
The goal is to build a predictive model of a maximum efficiency of new pump design as a function of nine input parameters such as blade angle, eye diameter, hub diameter, pump speed, number of impeller blades, etc. 
For the new pump design we have $18$ performance data and different setting of input parameters. 
We also have data available from five legacy pumps, with number of data varying between $10$ to $55$.
We used all these data to build a multi-source legacy data model for the new design. 
We carried out $10$-fold cross validation to demonstrate the improvement in accuracy of using multi-source legacy data modeling over other models.
Fig.~\ref{fig:pump_legacy} shows the cross-validation error against standard error of cross-validation for different models. 
The ``New-Sys" is the model built on new system data only, ``Legacy-k"  (for $k \in 1, \dots, 5$) are model based on multi-fidelity modeling by combining $k^{th}$ pump legacy data with new system data, and ``MS-Legacy" is the model based on multi-source legacy modeling. 
As seen, using multi-fidelity modeling using data from legacy pump-3 or pump-4 improves the model prediction when compared to model just build on new system data. 
However, using multi-source modeling using all the legacy data, improves the model significantly.

\subsection{Parallelizing GEBHM's MCMC}
\label{sec:asmc_gebhm}
While the GP does not assume a functional form of the problem, it is defined by a set of parameters, called hyperparameters, that need to be learned from the data.
The hyperparameters define the characteristics of the objective function, such as smoothness, magnitude, periodicity, etc. Accurately estimating these hyperparameters is a key ingredient in developing a reliable and generalizable surrogate model. 
Markov chain Monte Carlo (MCMC) is a ubiquitously used Bayesian method to estimate these hyperparameters. 
GEBHM is very effective on problems of small and medium size, typically less than 1000 training points.
However, traditional GP models do not scale well with respect to dimension and size of the dataset.
For some challenging industry applications, the predictive capability of the GP is required but each second during the training of the GP costs thousands of dollars.  
In this work, we apply a scalable MCMC-based methodology enabling the modeling of large-scale industry problems. 
To this end, we extend and implement in GEBHM an Adaptive Sequential Monte Carlo (ASMC) methodology for training the GP. This implementation saves computational time (especially for large-scale problems) while not sacrificing predictability over the current MCMC implementation. 
For more details on the theoretical workings on the method we refer interested readers to~\cite{pandita2019towards, bilionis2015crop}.
We demonstrate the effectiveness and accuracy of GEBHM with ASMC on on a two-objective challenging industry application in \sref{asmc_gebhm_app}.

\subsubsection*{Application: Parallelizing GEBHM's MCMC}
\label{sec:asmc_gebhm_app}
The problem is a GE combustion test data problem, where we aim to build a GP model to predict two emission quantities as a function of three measured temperatures, three fuel flow parameters, and air flow. Thus, there are nine hyperparameters that need to be estimated for each individual objective. 
The total number of hyperparameters for the GP model is 19. 
The MCMC settings for this problem are: 10000 steps with 1000 steps during the initialization.

 \begin{figure}[h]
  \subfigure[]{\includegraphics[width=0.5\textwidth]{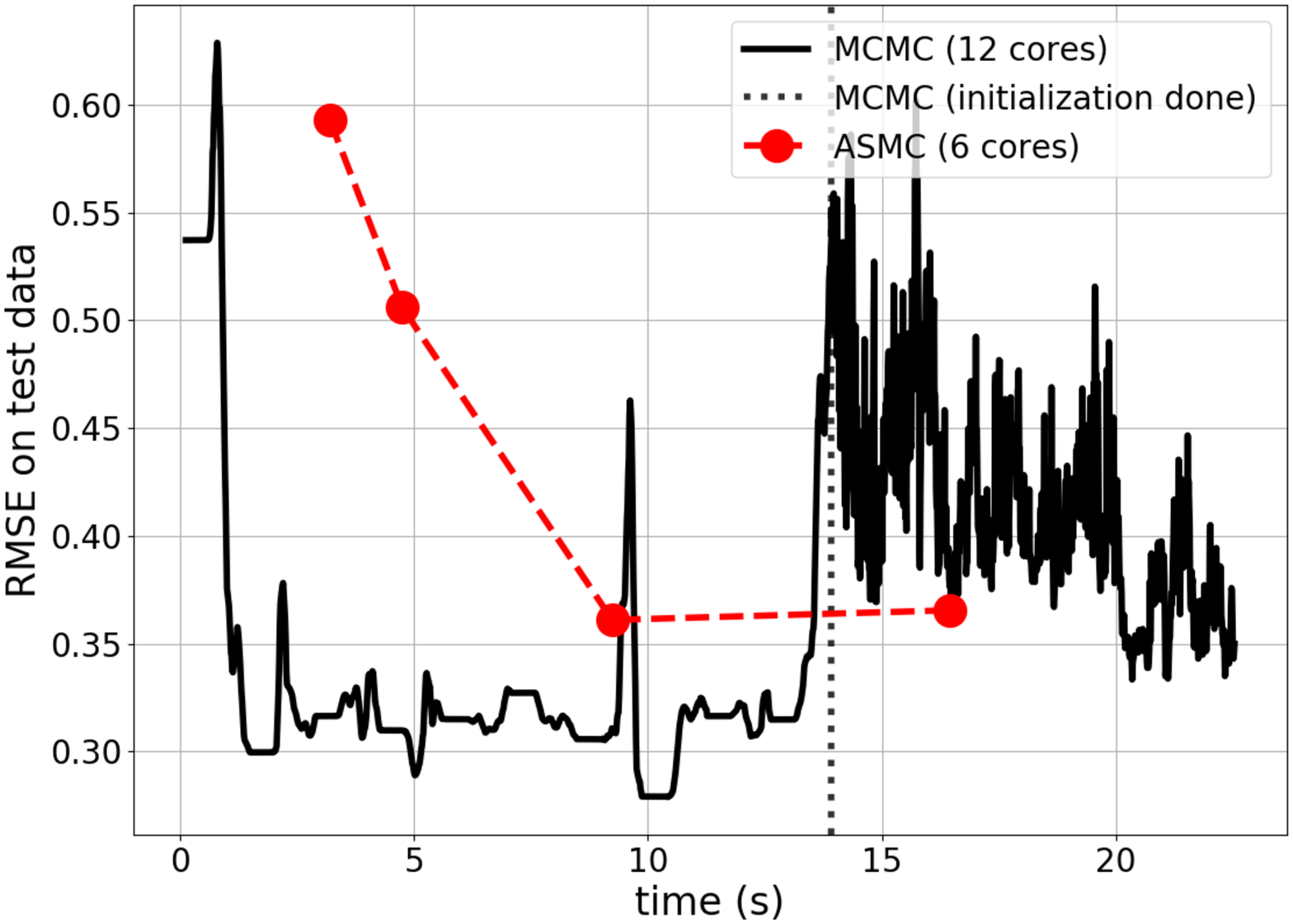}}
  \subfigure[]{\includegraphics[width=0.5\textwidth]{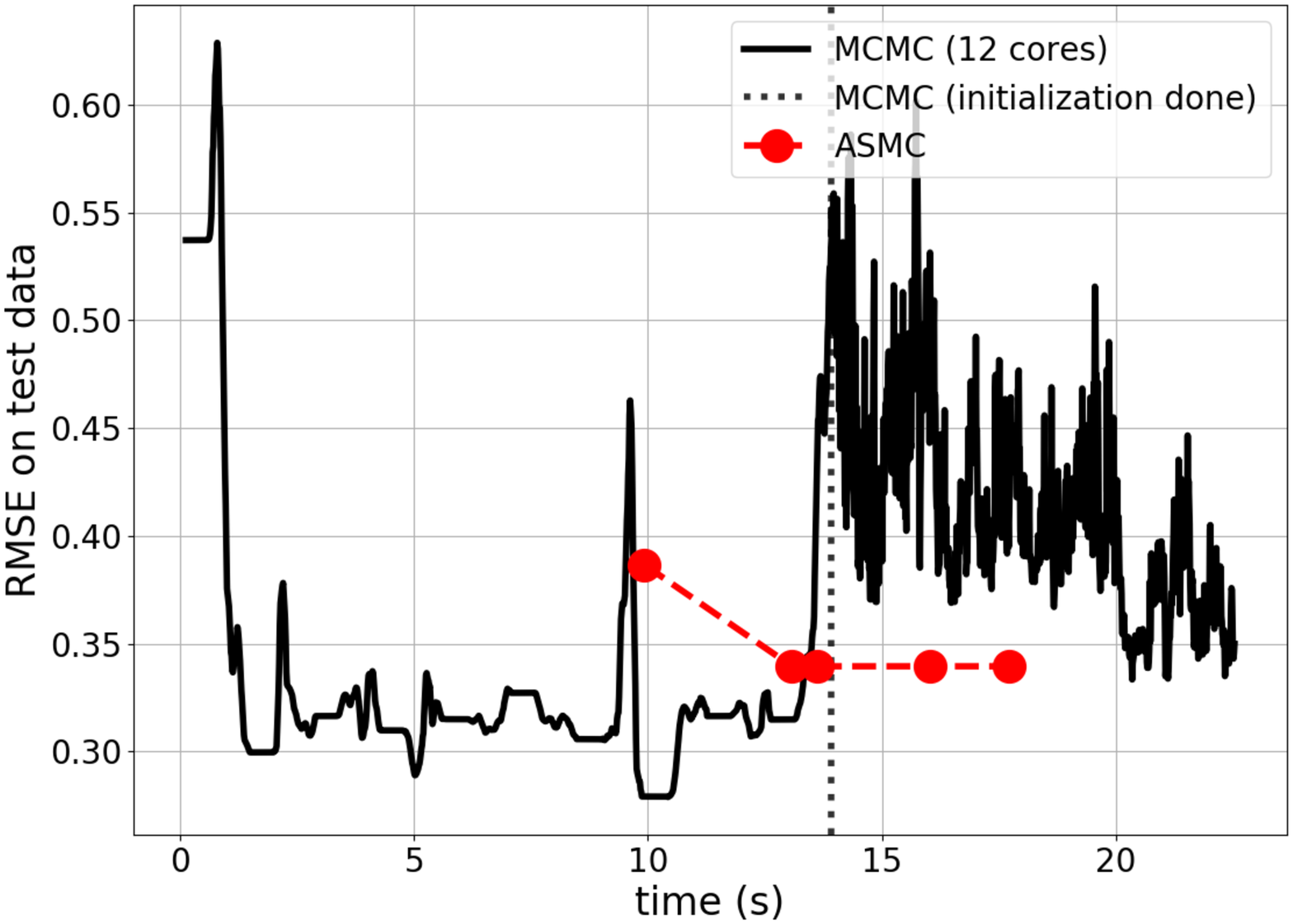}}
  \caption{Subfigure~(a) Number of particles on a workstation (red dots) are 6, 12, 30, and 60, and Subfigure~(b) number of particles on the HPC (red dots) are 24, 48, 120, 240, and 480, across the time axis, respectively for the first objective of the combustion problem. From \cite{pandita2019towards}}
  \label{fig:combustions_1}
 \end{figure}

 \begin{figure}[h]
  \subfigure[]{\includegraphics[width=0.5\textwidth]{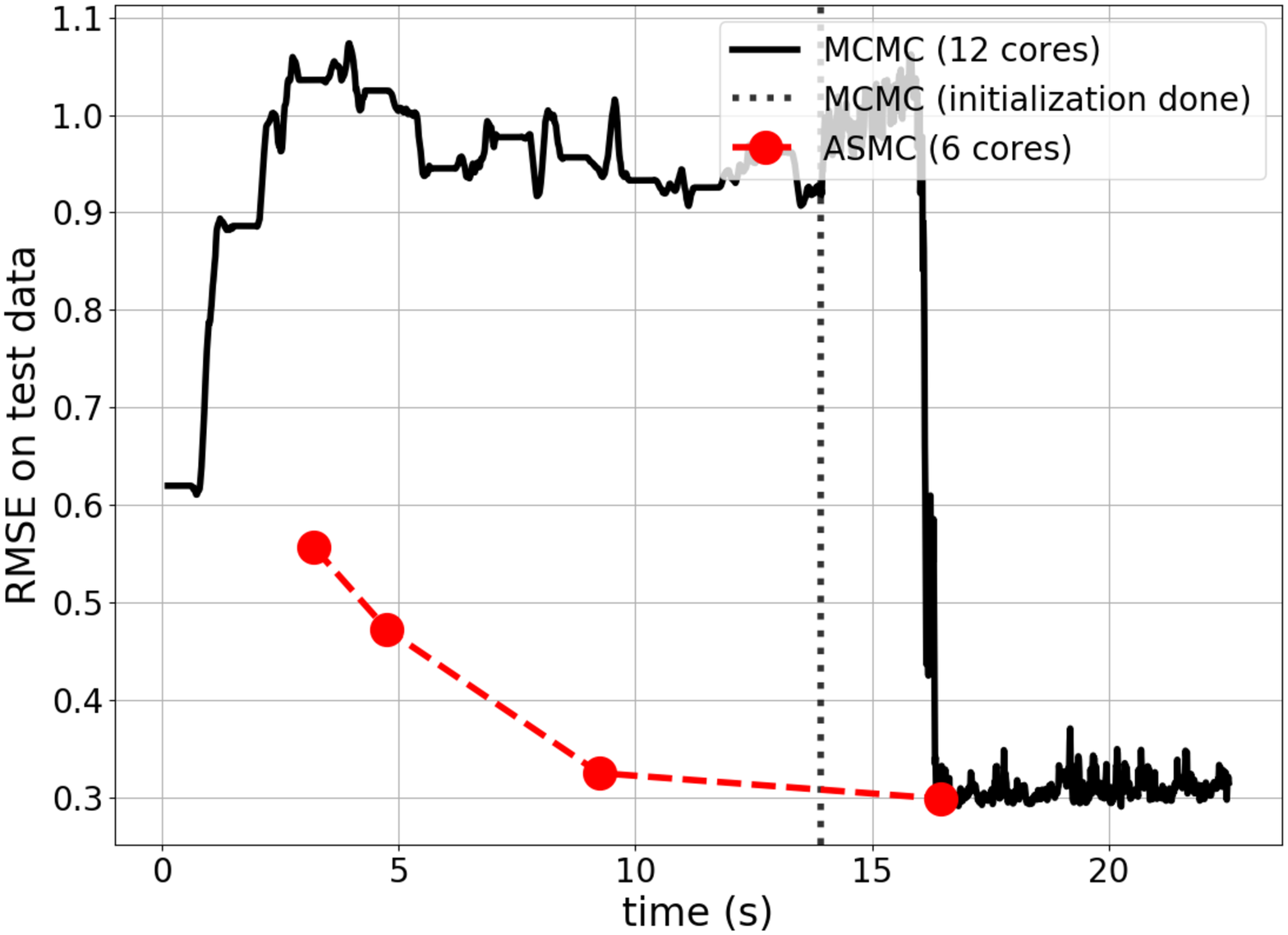}}
  \subfigure[]{\includegraphics[width=0.5\textwidth]{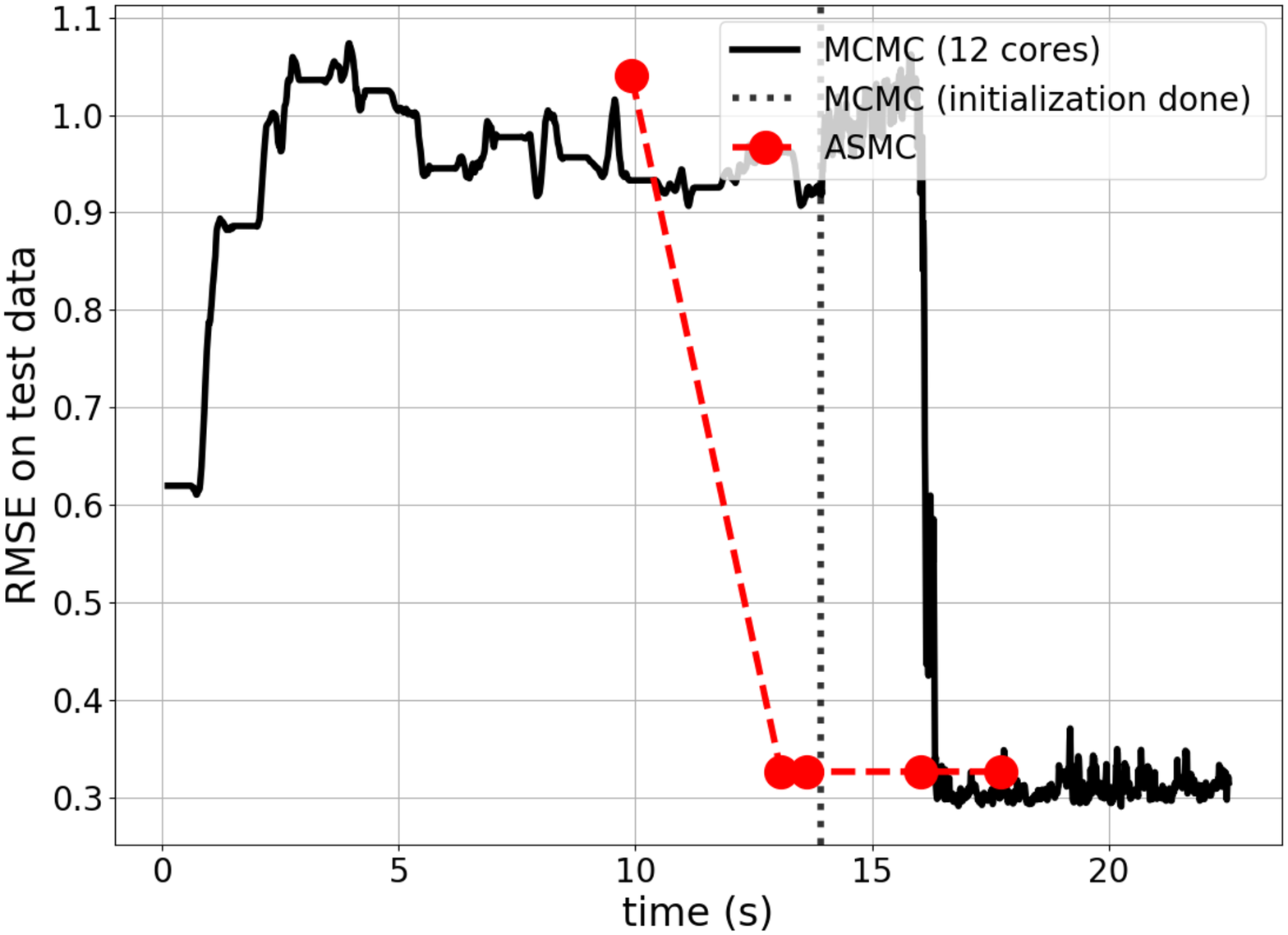}}
  \caption{Subfigure~(a) Number of particles on a workstation (red dots) are 6, 12, 30, and 60, and Subfigure~(b) number of particles on the HPC (red dots) are 24, 48, 120, 240, and 480, across the time axis, respectively for the second objective of the combustion problem. From \cite{pandita2019towards} }
  \label{fig:combustions_2}
 \end{figure}

The significance of the results in \fref{combustions_1} and \fref{combustions_2} is noticeable from the perspective of the number of objectives. Not only does the scaled-up HPC ASMC do well in terms of computational time but it improves the predictive accuracy of the GP model compared to GEBHM's MCMC. 
The number of steps per ASMC particle is one for the workstation and five for the HPC runs. 
The number of sequential distributions has been fixed to 10 for both the computing environments.

This provides a holistic overview of ASMC's performance on different types of problems. Problems of varying input dimensionality, varying number of objectives, and different training data sizes have been treated with the extended ASMC and compared to the current MCMC implementation. The ASMC does equally well or better compared to the MCMC in terms of predictive accuracy, while saving time by half or more in most of the challenging problems.

\section{Conclusion and future direction}
\label{Future}
We discussed GE's Bayesian probabilistic modeling framework, GEBHM, and various advanced methods developed for industrial applications. Although we only discussed the core capabilities of GEBHM in this work, these capabilities have also enabled development on other frontiers such as intelligent/adaptive design optimization ~\cite{kristensen2019industrial, Ling2018,kristensen2016}. The calibration and sensitivity analysis capabilities of GEBHM are demonstrated on a new  challenging industrial problem of structural dynamics as discussed. A summary of the numerous enhancements to the GEBHM framework are provided in~Table (\ref{summary_enhancement}). With the increase in complexity of industrial applications and demand of further improvement in probabilistic modeling methods,  some future directions of enhancement are as follows:

\begin{table}[h!]
	\small 
	\centering
	
\begin{tabular}{|p{3.5cm}|p{4cm}|p{3.5cm}|p{4cm}|} 
		 \hline
		 \textbf{Enhancement} & \textbf{Capability enabled} & \textbf{Current deficiency} & \textbf{Demonstrated application}\\ [0.5ex] 
		 \hline\hline
		 Transient problem & Handling of temporal responses & None &  1-D rod heat transfer \\ 
		 \hline
		 Robust Gaussian Process & Handling input uncertainty  & Handles only Gaussian noise on inputs & Airplane flight controller (NASA UQ challenge 2014)  \\
		 \hline
		 Bayesian sensitivity with correlated inputs & Handling sensitivity for correlated inputs & Handles only linear correlations &  Structural vibration problem\\
		 \hline		 
		 Portable Gaussian Process & Parametric representation of GEBHM meta-modeling & Limited predictive uncertainty representation & Crack growth problem \\
		 \hline		 
		 Multi-Source legacy modeling & Integrating multiple sources of data & Input dimensionality need to be same for all & Pump design problem \\
		 \hline		 
		 Parallelizing GEBHM’s MCMC & Enabling GEBHM to handle large scale industrial problems & Slower than deterministic approaches & Combustion test problem \\[1ex] 
		 \hline
	\end{tabular}

	\caption {Enhancements on GEBHM's core capabilities.}
	\label{summary_enhancement}
\end{table}

\subsection*{Preconditioned Conjugate Gradient Method}

The demanding computational cost and memory requirement constitute a significant challenge for GP to train and predict based on large datasets. The complexity stems from the requirement of multiple solutions of a large-scale system of linear equations involving the kernel matrix. Inspired by the concepts of divide-and-conquer, we will develop a scalable GP algorithm based on domain decomposition methods. The proposed algorithm will partition the hard-to-solve problem into smaller easy-to-solve ones which can be tackled concurrently. The solutions of the local problems will be utilized to construct a scalable preconditioner for the preconditioned conjugate gradient method (PCGM) to tackle the global problem. For complex industrial applications, a multi-level and multi-fidelity preconditioner will be essential for cost-effective performance and robustness of the algorithm~\cite{subber2014schwarz,subber2013dual,subber2014domain}.

\subsection*{Scalable Gaussian process}

With the rapid development of high-performance computing, Big Data and Internet of Things, it has become necessary to scale up GPs for modeling large dataset. Various frameworks have been proposed to scale up GPs such as sparse Gaussian Process for global approximation and distributed Gaussian Process for local approximation. Most existing schemes focus on speeding up the training while balancing the overall prediction accuracy. However, there are certain cases requiring special treatment. For example, the dataset could be from non-uniform distribution in which the dataset are scattered as clusters among the input space. For modeling large datasets with non-uniform distributions, the scalable schemes might overlook the clusters with relatively less data. This will deteriorate the predicative performance at the overlooked area. It's possible to fix the issue with intelligent selection of a subset for GP development with a reasonable coverage among the whole dataset ~\cite{zhang2020remarks}. The intelligent selection should take into account both input variables and the output values. A small subset is expected to be adequate while assuming the large dataset contains more-than-enough data for model development.

\subsection*{Deep Gaussian processes}
Gaussian process models have remarkable data efficiency properties due to the very narrow, yet reasonable, prior beliefs that they place on the function being learned.
A core element of this prior is the GP kernel, which encodes beliefs about the regularity and differentiability of the function being modeled.
However, traditional kernels are often selected out of convenience rather than out of the belief that they accurately reflect one's prior beliefs.
There are many systems that are known to violate familiar assumptions such as stationarity, homoscedasticity, and Gaussian marginal densities.
One way of overcoming these limitations is by learning new representations of the data being modeled to enter as inputs to the Gaussian process simultaneously with training the GP itself.
In this spirit, recent model architectures including deep kernel learning \cite{wilson2016deep} and deep Gaussian processes \cite{damianou2013deep} are have been shown to improve predictive performance and are becoming increasingly accessible due to the advent of advanced approximate inference techniques \cite{titsias2009variational, titsias2014doubly, salimbeni2017doubly}, software platforms \cite{abadi2016tensorflow, paszke2017automatic, matthews2017gpflow}, and modern computing hardware.
We see considerable opportunity as the community continues to experiment with combining these techniques with traditional modeling approaches.

\subsection*{Modeling multiple datasets}
Up to this point, our discussion has focused mainly on modeling a single input-output relationship.
However, there are many general classes of problems inside which small permutations alter the details of the input-output relationship being modeled.
One could imagine that the similar problems previously solved might be used to inform the model for the problem at hand.
We are investigating the use of compositional models in which task labels are embedded as latent variables in multi-dimensional Euclidean space; these latent variables augment the input space, allowing one to simultaneously learn a single model over the augmented input space while learning an embedding of tasks that enables a parsimonious functional representation.
We recognize that while one may assemble very large datasets by combining tasks, the complexity of the input space will grow at the same time.
Thus, there remains a strong case to be made for incorporating uncertainty associated with both the learned embedding as well as the function relating the augmented input space to the outputs.
By following the Bayesian modeling groundwork laid out in the previous sections, we anticipate that one will be able to learn credible models that can be trained with even fewer data than what the BHM framework enables today.

\section*{Acknowledgments}
The authors acknowledge the contributions from Dr. Ankur Srivastava, Dr. Felipe Viana, Dr. Arun K. Subramaniyan, Dr. Issac Asher, Dr. You Ling, Dr. Kevin M. Ryan, and Dr. Jesper Kristensen for their critical contribution in development and enhancement of GEBHM. 

\bibliographystyle{asmems4}

\bibliography{asme2e}

\end{document}